\documentclass[10pt, a4paper]{article}

\usepackage{lrec-coling2024} % this is the new style
\usepackage{booktabs} 
\usepackage{multirow} 

\title{HAE-RAE Bench: \\ Evaluation of Korean Knowledge in Language Models}

\name{Guijin Son, Hanwool Lee, Suwan Kim, Huiseo Kim, Jaecheol Lee \\
\textbf{ \large Je Won Yeom, Jihyu Jung, Jung Woo Kim, and Songseong Kim}}

\address{EleutherAI, OnelineAI, MODULABS \\
         spthsrbwls123@yonsei.ac.kr}

\abstract{
Large language models (LLMs) trained on massive corpora demonstrate impressive capabilities in a wide range of tasks. While there are ongoing efforts to adapt these models to languages beyond English, the attention given to their evaluation methodologies remains limited. Current multilingual benchmarks often rely on back translations or re-implementations of English tests, limiting their capacity to capture unique cultural and linguistic nuances. To bridge this gap for the Korean language, we introduce the \textbf{HAE-RAE Bench}, a dataset curated to challenge models lacking Korean cultural and contextual depth. The dataset encompasses six downstream tasks across four domains: vocabulary, history, general knowledge, and reading comprehension. Unlike traditional evaluation suites focused on token and sequence classification or mathematical and logical reasoning, the HAE-RAE Bench emphasizes a model's aptitude for recalling Korean-specific knowledge and cultural contexts. Comparative analysis with prior Korean benchmarks indicates that the HAE-RAE Bench presents a greater challenge to non-Korean models by disturbing abilities and knowledge learned from English being transferred. 
 \\ \newline  }

\begin{document}

\maketitleabstract

\section{Introduction}

Over time, LLMs and benchmark datasets have evolved in tandem, continually becoming more sophisticated and challenging, recognizing their reciprocal relationship. Despite the pivotal role played by benchmark datasets in advancing the capabilities of LLMs, multilingual evaluation tools remain primarily limited. Existing evaluation efforts often rely on translated versions of English datasets~\citep{shi2022language} or translation-specific benchmarks such as WMT 21~\citep{akhbardeh-etal-2021-findings}. While providing some insights into the models' performance across languages, this approach fails to fully capture the intricacies, nuances, and knowledge specific to each linguistic context.

Some of the existing efforts to evaluate language models in Korean include Korean-NLI \& STS~\citep{ham-etal-2020-kornli}, KLUE~\citep{park2021klue}, and KoBEST~\citep{kim2022kobest}. Korean-NLI \& STS is derived from machine and human translations of English datasets for natural language inference (NLI) and semantic textual similarity (STS). Accordingly, they hardly capture the unique nuances of the Korean language. KLUE is a Korean version of the GLUE benchmark~\citep{wang2018glue}, which supports a variety of tasks, including NLI, STS, and topic classification. Unfortunately, its adoption was limited due to its relatively simple tasks. The latest benchmark, KoBEST, is designed to assess a language model's ability to address questions that require advanced reasoning, like understanding passages of time or causality. However, with the advent of Large Language Models (LLMs) such as GPT-4~\citep{OpenAI2023GPT4TR} and conversational agents built upon them, there is an increasing need to evaluate the cultural knowledge of language models to ensure they converse with native speakers without sounding incoherent. To address this issue, we introduce the \textbf{HAE-RAE Bench}, a Korean benchmark dataset originally crafted to capture culture-specific nuances inherent to the Korean language. 

\begin{figure}[t]
\includegraphics[width=\columnwidth]{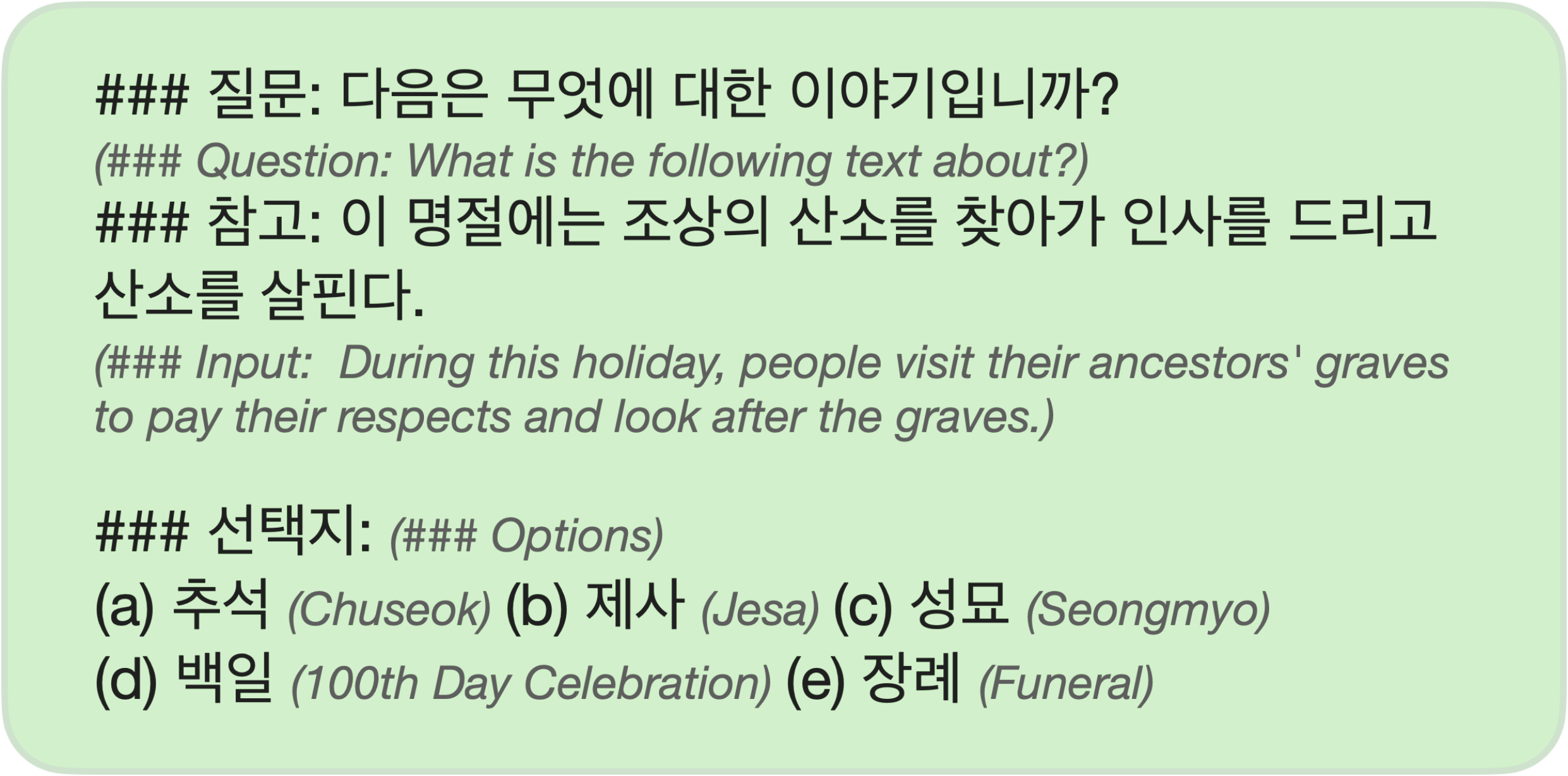}
\centering
\caption{\footnotesize Example instance from the HAE-RAE Bench. English translations are added for broader accessibility.}
\label{kgkd}
\end{figure}

We evaluate ten language models including, Polyglot-Ko~\citep{ko2023technical}, UMT5~\citep{chung2023unimax}, Llama-2~\citep{touvron2023llama}, GPT-3.5-Turbo and GPT-4~\citep{OpenAI2023GPT4TR}. Evaluation results reveal that multilingual LLMs suffer in solving the HAE-RAE Bench compared to Polyglot-Ko, a native language model trained on Korean from scratch. Furthermore, we our results hint that In-Context Learning (ICL) may be insufficient in steering a LLM to align with a specific culture. HAE-RAE Bench is publicly available for future research.\footnote{\url{https://huggingface.co/datasets/HAERAE-HUB/HAE_RAE_BENCH}}

\section{Related Work}
\subsection{Language Model}

Since the introduction of the Transformer architecture~\citep{vaswani2017attention} and early derivatives like BERT~\citep{devlin2018bert} and GPT~\citep{radford2018improving}, research in English language models has expanded rapidly. With their instruction-following capabilities, InstructGPT~\citep{instructgpt} and Flan-T5~\citep{flant5} further invigorated this interest. While most of these models primarily focus on English, there are notable exceptions for Chinese, with Qwen~\citep{qwen}, Baichuan~\citep{yang2023baichuan}, and GLM~\citep{zeng2022glm}. Efforts aimed at narrowing the disparity in progress between English and other languages include:
\begin{enumerate}
    \item Building language-specific models from scratch, such as Polyglot-Ko(Korean)~\citep{ko2023technical}, HyperCLOVA(Korean)~\citep{kim-etal-2021-changes}, Japanese StableLM(Japanese)~\citep{jlm}, and ruGPT(Russian)~\citep{rugpt};
    \item Developing multilingual models like BLOOM~\citep{scao2022bloom}, MT5~\citep{xue2020mt5}, and UMT5~\citep{chung2023unimax};
    \item Adapting English models for other languages, as seen with Sabiá~\citep{pires2023sabi} and Chinese-LLaMA~\citep{cui2023efficient}.
\end{enumerate}

Following the advancement of multilingual language models, a critical research question arises: "How should the language-specific capabilities of these models be evaluated?" This underscores the necessity for benchmarks specifically curated to assess the multilingual ability of LLMs.

\subsection{Multilingual Evaluation}

Multitask benchmarks like GLUE~\citep{wang2018glue} and SuperGLUE~\citep{wang2019superglue} were introduced along with the English language models. Once these were saturated, they were followed by even bigger benchmarks such as MMLU~\citep{hendrycks2020mmlu} and Big BENCH~\citep{srivastava2022big}. Non-English evaluation research has mirrored this trend, predominantly through translation or re-implementation of existing English benchmarks. Examples include JGLUE~\citep{kurihara-etal-2022-jglue}, KLUE~\citep{park2021klue}, and CMMLU~\citep{li2023cmmlu}, which are Japanese and Korean adaptations of GLUE and Chinese re-implementation of MMLU, respectively.

However, these benchmarks fall short of capturing the native knowledge encoded in the parameter of LLMs. This highlights the need for evaluation suites curated to assess the cultural context of a model. Recent research in this direction is BHASA~\citep{leong2023bhasa}, which aims at gauging the cultural depth of language models in Southeast Asian languages. Nonetheless, limitations are apparent: only 34 questions for Indonesian and 28 for Tamil in the entire dataset specifically address cultural representation tasks. In this paper, we introduce the \textbf{HAE-RAE Bench}, an evaluation set of 1.5K questions curated to assess Korean-specific knowledge in language models.

\subsection{Korean Evaluation}

Korean language model evaluation is also a field of interest, with resources emerging after English and Chinese. Examples include Korean-NLI \& STS~\citep{ham-etal-2020-kornli}, KorFin-ASC~\citep{son2023removing}, KLUE~\citep{park2021klue}, and KoBEST~\citep{kim2022kobest}. Korean-NLI \& STS are based on translations of English datasets for natural language inference (NLI) and semantic textual similarity (STS), potentially missing Korean nuances. KorFin-ASC is derived from Korean news but concentrates on sentiment classification, specifically in the financial domain. KLUE mirrors the GLUE benchmark with Korean, covering tasks like Topic Classification, Semantic Textual Similarity, Natural Language Inference, and more. However, either translated or task-oriented, these benchmarks fail to fully assess language-specific models. Recent research include KoBEST~\citep{kim2022kobest}, which features Korean re-implementations of HellaSwag~\citep{zellers2019hellaswag}, COPA~\citep{gordon-etal-2012-semeval}, BOOLQ~\citep{clark-etal-2019-boolq}, SentiNeg~\citep{sentineg}, and WiC~\citep{pilehvar2018wic}. However, LLMs trained in massive amounts of English corpora may excel in these evaluation suites by leveraging their general problem-solving capabilities derived from the scale. The HAE-RAE Bench distinguishes itself from the above-mentioned Korean benchmarks by evaluating the depth of knowledge encoded in language models instead of their natural language understanding or reasoning abilities.

\section{HAE-RAE Bench}

\begin{table*}[htbp]
\centering
\fontsize{8}{11}\selectfont
\begin{tabular}{lccccccc}
\toprule
 & & \multicolumn{2}{c}{\textbf{Total \# of}} & \multicolumn{2}{c}{\textbf{Avg. \# of Words (std)}} &  \multicolumn{2}{c}{\textbf{Fertility Rate (std)}} \\ \cmidrule(lr){3-4} \cmidrule(lr){5-6} \cmidrule(lr){7-8}
\multicolumn{1}{c}{Category} & Type & Question & Unique Morpheme &  per question & per passage &  Polyglot-Ko & Llama-2 \\ \midrule
Loan Words & \{Q\} & 169 & 960 &  5.1 (0.3) & - & 3.9 (0.3) &  6.7 (0.6) \\
Rare Words & \{Q\} & 405 & 2721 & 13.0 (3.4) & - &  3.1 (0.3)  & 6.1 (0.4) \\
Standard Nomenclature & \{Q\} & 153 & 1018 &  8.3 (0.5) & - &  3.2 (0.4)  & 6.4 (0.6) \\
Reading Comprehension & \{Q, P\} & 447 & 5825 & 7.1 (1.8) & 69.6 (44.6) & 2.5 (0.4) & 6.0 (0.5) \\
General Knowledge & \{Q, P\} & 176 & 2099  & 7.0 (3.0) & 9.1 (13.6)  & 3.4 (0.6) &  6.4 (0.9) \\
History & \{Q\} & 188 & 1595 & 12.8 (3.5) & -  & 3.3 (0.4) & 6.3 (0.6) \\ \bottomrule
\end{tabular}
\caption{\footnotesize HAE-RAE Bench Statistics.}
\label{hrstat}
\end{table*}

The design principle behind the HAE-RAE Bench significantly differs from earlier Korean benchmark suites like KLUE~\citep{park2021klue} or KoBEST~\citep{kim2022kobest}. While previous benchmarks focused on evaluating natural language understanding or reasoning abilities, HAE-RAE emphasizes the depth of knowledge itself. This change is driven by the emergence of LLMs and conversational agents or search engines built on them. We posit that knowledge of Korean vocabulary, culture, geography, and history might be as crucial, if not more so, than traditional NLU tasks such as token or sequence classification in conversational situations. Accordingly, the resulting benchmark encompasses six downstream tasks: Loan Words(LW),  Standard Nomenclature(SN), Rare Words(RW), General Knowledge(GK), History(HI) and Reading Comprehension(RC). 

Statistics for the HAE-RAE Bench dataset are provided in Table~\ref{hrstat}. ``Type" indicates the structure of the question. ``Q" denotes that the instance comprises a question with multiple choices, while ``Q, P" indicates the inclusion of an associated passage. We also present the fertility rate of the dataset, tokenized using different models: Polyglot-Ko~\citep{ko2023technical}, UMT5~\citep{chung2023unimax}, and Llama-2~\citep{touvron2023llama}. The fertility rate~\citep{fertility} calculates the average number of sub-tokens generated per word. A fertility rate of 1 implies that the tokenizer's vocabulary encompasses every word in the text. A higher fertility rate may suggest potential challenges for the tokenizer in grasping context. In our observation, the fertility rate increases for models with less emphasis on Koreans. To assess the relative complexity of the vocabularies in the HAE-RAE Bench, we compared its fertility rate with that of KoBEST, as shown in Table~\ref{fer-comp}. Using the polyglot-ko tokenizer, the fertility rates for HAE-RAE Bench and KoBEST are 3.0 and 2.7, respectively. This suggests that the HAE-RAE Bench comprises less common words. Examples for each subset of the datasets are presented in section~\ref{a:ex}.

\begin{table}[htbp]
\centering
\fontsize{7.5}{11}\selectfont
\begin{tabular}{lccc}
\hline
\textbf{Dataset} & \textbf{Polyglot-Ko} & \textbf{UMT5} & \textbf{Llama-2} \\ \hline
HAE-RAE Bench & 3.00 (0.38) & 3.56 (0.37) & 6.33 (0.59) \\
KoBEST & 2.70 (0.34) & 3.39 (0.45) & 6.44 (0.75) \\ \hline
\end{tabular}%
\caption{\footnotesize Fertility rate (std) of HAE-RAE Bench and KoBEST.}
\label{fer-comp}
\end{table}

\subsection{Loan Words\label{LW}}
\paragraph{Task Description}~Loan words refer to vocabularies directly adopted from foreign languages. In South Korea, the National Institute of Korean Language (NIKL)~\footnote{\url{https://www.korean.go.kr/front/main.do}} formulates corresponding Korean terms for such words. In this task, language models are given a foreign word along with five choices and are tasked to identify the correct Korean equivalent.

\paragraph{Creation} The pairs of foreign words and their Korean equivalents are sourced from NIKL. Some Korean terms are infrequently used, either because the foreign word has been entrenched in society for a long time or because it's a recent addition and not yet widely recognized. To ensure we focus on reasonably common terms, we filter the list to only include words present in both ``Naver Knowledge Encyclopedia"~\footnote{\url{https://terms.naver.com}} and ``Daum Encyclopedia"~\footnote{\url{https://100.daum.net}}, the two most widely used online encyclopedias in Korea. From the refined list, we randomly sampled 200 vocabularies. Incorrect options were selected from the remaining terms based on their Levenshtein distance~\citep{levenshtein1966binary} to the correct answer. While Levenshtein distance may initially seem to prioritize syntax over semantics, it effectively captures both in Korean. ``Han" (Chinese logograms) constitute about 55\% of the Korean vocabulary. Accordingly, words with the same Korean letter have related meanings. Moreover, the structure of the Korean language involves compounding, where multiple ``roots" (fundamental word units) merge to form new words. Consequently, words sharing similar meanings often include the same ``root", making the syntactic and semantic distances in Korean words largely aligned. Finally, we applied a Levenshtein distance threshold of 3, omitting samples with fewer than four incorrect options meeting this criterion.

\subsection{Standard Nomenclature}
\paragraph{Task Description} Standard Nomenclatures, published by NIKL, are unified terminology for domain-specific words. In this task, language models are presented with a specialized term along with five options, with the objective of identifying the official term endorsed by NIKL. 

\paragraph{Creation} Pairs of domain-specific words and their official terms are collected from NIKL. We follow the approach in \ref{LW} to create questions.

\subsection{Rare Words}
\paragraph{Task Description} The Rare Words task aims to probe language models' understanding of challenging vocabulary. Given a definition and five words, models are tasked with selecting the word that best suits the provided definition. 

\paragraph{Creation} We sourced pairs of definitions and challenging words from past episodes of the TV program ``Woorimal Battle,"~\footnote{\url{https://program.kbs.co.kr/1tv/culture/woorimal/pc/index.html}} known for its challenging Korean vocabulary quizzes. We follow the approach in \ref{LW} to create questions.

\subsection{General Knowledge}
\paragraph{Task Description} General Knowledge evaluates the model's familiarity with various aspects of the Korean cultural, using five-option multiple-choice questions.

\begin{table}[h]
\centering
\fontsize{8}{11}\selectfont
\begin{tabular}{lcc}
\toprule
\textbf{Category}     & \textbf{\# of instances} & \textbf{Average Length} \\ \midrule
Tradition      & 17        & 35.2           \\
Law          & 10        & 32.2           \\
Geography    & 49        & 46             \\
Korean Pop   & 50        & 42.3           \\
Korean Drama & 50        & 36.7           \\  \bottomrule
\end{tabular}%
\caption{\footnotesize The number of data instances for each category.}
\label{gk-sub}
\end{table}

\paragraph{Creation} We first identify five primary categories for general knowledge: tradition, law, geography, Korean pop, and Korean drama. We then crowd-sourced questions to fit these subcategories. Then, we remove overlapping, factually incorrect, and questions that fail to align with the defined category. We also conduct additional investigations to ensure that no superficial artifacts are not inadvertently introduced. Basic statistics for each subcategory are illustrated in  Table~\ref{gk-sub}.

\paragraph{Investigation} Following \citet{kaushik2018much}, we examined the performance of Polyglot-Ko-12.8B using question-only (Q-only) and context-only (C-only) settings. Polyglot-Ko-12.8B achieved scores of 25.57\%  and 23.86\%  for Q-only and C-only, respectively, while the complete setting outperformed both with a score of 32.95\%. Although the Q-only and C-only settings are within 10\% accuracy of the complete setting, it is worth noting that the model's lower bound is set at 20\%. Therefore, we conclude that the dataset was crafted correctly to require both question and context to answer.

\begin{table}[ht]
\centering
\fontsize{8}{11}\selectfont
\begin{tabular}{ccccc}
\toprule
\textbf{Metric}  & \textbf{Full}  & \textbf{Q-Only} & \textbf{C-Only} & \textbf{$\Delta$ (\textit{min})} \\ \midrule
Acc & \textbf{32.95} & 25.57  & 23.86 & -7.38   \\ 
Macro F1 & \textbf{32.01} & 24.35  & 23.64 & -7.56   \\ 
\bottomrule
\end{tabular}%
\caption{\footnotesize Performance of Polyglot-Ko-12.8B on General Knowledge with truncated inputs.}
\label{gk-invest}
\end{table}

\subsection{History}
\paragraph{Task Description}  The history task assesses the model's understanding of historical events. Presented with a question and five options, the model must identify the correct answer. 

\paragraph{Creation} We first sourced web pages tagged ``Korean history" from Namuwiki, Korea's equivalent to Wikipedia, and randomly selected 40 pages. From each page, authors manually crafted five questions. We refer \citet{malaviya-etal-2022-cascading} and filtered out 12 questions with overlapping tokens between questions and answers. Moreover, to investigate potential biases introduced while creating the wrong options, we analyzed two simple linguistic indicators: the probability of the longest option being correct was 21.53\%, and for the shortest option, it was 17.01\%. Through this process, we remove overly simplistic questions and investigate for potential biases.

\subsection{Reading Comprehension}
\paragraph{Task Description} Reading comprehension tasks involve providing paired questions and passages along with four options. The materials for our Reading Comprehension (RC) tests were sourced from the Korean Language Ability Test (KLAT),  an exam designed to evaluate proficiency in Korean as a second language. 

\paragraph{Creation} The tests were gathered from sample materials publicly released by the Korea Educational Testing Service (KETS). We omitted questions that required interpreting images. The sourced KLAT is divided into four proficiency tiers: three that correspond to the Common European Framework of Reference (CEFR) levels—A (beginner), B (intermediate), and C (advanced)—plus an introductory level below A for absolute beginners.

\subsection{Quality Check} 
To further filter the collected questions, we reviewed the entire dataset and conducted factual verification using online resources. In this process, we manually corrected 23 questions with labeling or crawling errors.

\section{Evaluation Settings}
\subsection{Language Models}
We evaluated ten models across varying sizes from four model families. From openly available models we selected (1) Korean-focused models: Polyglot-ko-1.3B/3.8B/5.8B/12.8B~\citep{ko2023technical}, (2) Multilingual models: UMT5-XL/XXL~\citep{chung2023unimax}, and (3) English-centric models: Llama-2-7B/13B~\citep{touvron2023llama}. For analysis, we excluded models that do not disclose statistics on the number of pretrained Korean tokens. This leaves out Falcon~\citep{penedo2023refinedweb} and BLOOM~\citep{scao2022bloom} from our experiments. Additionally, we included GPT-3.5-Turbo and GPT-4 in our evaluation to gauge the efficacy of the HAE-RAE Bench in assessing state-of-the-art proprietary LLMs.

\paragraph{Polyglot-Ko} \citet{ko2023technical} is available in four sizes: 1.3B, 3.8B, 5.8B, and 12.8B, all built using the GPT-NeoX codebase. It was pretrained on a Korean-only corpus, with sizes ranging between 167B to 212B tokens. Despite its smaller pretraining budget compared to similar-sized English models, Polyglot-Ko achieved state-of-the-art results on KoBEST, a benchmark comprising five Korean language understanding and reasoning tasks~\citep{kim2022kobest}.

\paragraph{UMT5} \citet{chung2023unimax} was originally trained in five sizes: small (77M), base (250M), large (800M), xlarge (3B), and xxlarge (13B), closely following the mT5 architecture\citep{xue2020mt5}. However, the large variant was not released publicly due to pretraining instability. The models are trained on a corpus of 1T tokens, which includes 14.8 billion Korean tokens. UMT5 surpasses mT5 in benchmarks such as XNLI~\citep{conneau2018xnli} and TyDi QA~\citep{clark-etal-2020-tydi}. As the small and base models do not have counterparts in the Polyglot-Ko suite, our experiments focus on the xlarge and xxlarge models.

\paragraph{Llama-2} \citet{touvron2023llama} is available in three sizes: 7B, 13B, and 70B. It is trained on a corpus of 2T tokens, predominantly in English (89.7\%), with Korean comprising a mere 0.06\% or about 0.6B tokens. We utilize the version without fine-tuning. The Llama-2-70B model is excluded from our study due to the absence of a corresponding Korean model.

\begin{table*}[htbp]
\centering
\fontsize{9}{11}\selectfont
\begin{tabular}{lcrcrrcrcrr}
\toprule
 & & \multicolumn{3}{c}{\textbf{Loan Words}} &  \multicolumn{3}{c}{\textbf{Standard Nomenclature}} & \multicolumn{3}{c}{\textbf{Rare Words}} \\ 
 \cmidrule(lr){3-5} \cmidrule(lr){6-8} \cmidrule(lr){9-11}
Model & \multicolumn{1}{l}{Params} & \multicolumn{1}{c}{n=0} & \multicolumn{1}{c}{n=5} & \multicolumn{1}{c}{n=10} &  \multicolumn{1}{c}{n=0} & \multicolumn{1}{c}{n=5} & \multicolumn{1}{c}{n=10} &  \multicolumn{1}{c}{n=0} & \multicolumn{1}{c}{n=5} & \multicolumn{1}{c}{n=10}  \\ \midrule
\multirow{4}{*}{Polyglot-Ko} & 1.3B & 76.92 & 88.76 & 91.72 &  60.13 & 69.93 & 71.24 &  47.41 & 61.48 & 61.23 \\
 & 3.8B & 78.70 & 88.76 & 91.72 & 63.40 & 79.74 & 77.78 &  47.16 & 70.62 & 72.10 \\
 & 5.8B & 82.84 & 93.49 & 94.08 &  \textbf{66.67} & 82.35 & 83.66 &  \textbf{56.79} & 73.09 & 74.57 \\
 & 12.8B & \textbf{87.57} & \textbf{94.67} & \textbf{94.67} &  61.44 & \textbf{84.97} & \textbf{86.93} &  53.09 & \textbf{75.31} & \textbf{76.05} \\ \midrule
\multirow{2}{*}{UMT5} & 3B & 58.58 & 61.54 & 59.76 &  41.83 & 37.25 & 33.33 &  25.68 & 25.43 & 24.44 \\
 & 13B & 58.58 & 59.76 & 60.36 &  41.83 & 43.79 & 44.44 &  33.09 & 30.37 & 28.64 \\\midrule
\multirow{2}{*}{LLaMA-2} & 7B & 66.86 & 73.96 & 75.15 &  39.22 & 49.02 & 50.98 &  29.38 & 39.26 & 39.01 \\
 & 13B & 66.86 & 77.51 & 78.11 & 49.02 & 57.52 & 64.05 &  32.35 & 42.47 & 43.95 \\ \bottomrule
\end{tabular}%
\caption{\footnotesize Evaluation results of the performance on Loan Words, Standard Nomenclature, and Rare Word tasks.}
\label{evresults1}
\end{table*}

\begin{table*}[htbp]
\centering
\fontsize{9}{11}\selectfont
\begin{tabular}{lrrrrrrrrrc}
\toprule
 &  & \multicolumn{3}{c}{\textbf{History}} & \multicolumn{3}{c}{\textbf{General Knowledge}} & \multicolumn{3}{c}{\textbf{Reading Comprehension}} \\ 
 \cmidrule(lr){3-5} \cmidrule(lr){6-8} \cmidrule(lr){9-11}
Model & \multicolumn{1}{l}{Params} & \multicolumn{1}{c}{n=0} & \multicolumn{1}{c}{n=5} & \multicolumn{1}{c}{n=10} &  \multicolumn{1}{c}{n=0} & \multicolumn{1}{c}{n=5} & \multicolumn{1}{c}{n=10} &  \multicolumn{1}{c}{n=0} & \multicolumn{1}{c}{n=5} & \multicolumn{1}{c}{n=10} \\ \midrule
\multirow{4}{*}{Polyglot-Ko} & 1.3B & 60.11 & 78.19 & 77.13 &  26.70 & 30.68 & 28.98 & 34.45 & 37.81 & 37.14 \\
 & 3.8B & 69.15 & 86.17 & 85.11 &  28.41 & \textbf{33.52} & 33.52 &  40.49 & 42.06 & 40.04 \\
 & 5.8B & 79.79 & 85.11 & 81.91 &  29.55 & 27.84 & 28.41 &  40.72 & 42.73 & 41.39 \\ 
 & 12.8B & \textbf{80.32} & \textbf{88.30} & \textbf{90.43} & \textbf{32.95} & \textbf{33.52} & \textbf{34.66} &  \textbf{41.61} & \textbf{45.41} & \textbf{46.76} \\ \midrule
\multirow{2}{*}{UMT5} & 3B & 14.36 & 12.77 & 14.36 &  22.73 & 19.32 & 19.32 &  25.28 & 24.83 & 25.28 \\
 & 13B & 21.59 & 18.09 & 19.15 &  21.81 & 25.00 & 19.32 &  29.75 & 25.28 & 27.74 \\ \midrule
\multirow{2}{*}{LLaMA-2} & 7B & 28.72 & 35.64 & 35.64 &  21.02 & 24.43 & 25.00 &  29.98 & 32.89 & 31.32 \\
 & 13B & 35.11 & 38.83 & 40.96 &  28.41 & 31.82 & 28.98 &  31.99 & 36.47 & 34.00 \\ \bottomrule
\end{tabular}%
\caption{\footnotesize Evaluation results of the performance on History, General Knowledge, and Reading Comprehension tasks.}
\label{evresults2}
\end{table*}

We employ the ``log-likelihood" method implemented via LM-Eval-Harness~\citep{eval-harness} to evaluate the models. We compute the log-likelihood with each option concatenated to the question and select the one with the highest likelihood as the answer. All evaluations are implemented using bfloat16 precision in 0-shot, 5-shot, and 10-shot settings. We use accuracy as our primary metric.

HAE-RAE Bench aims to curate a dataset that challenges models lacking depth in Korean culture and knowledge, thereby guiding researchers in creating better Korean language models. To compare the ability of this benchmark to differentiate less native language models against prior benchmarks, we use KoBEST~\citep{kim2022kobest} as our baseline. We selected KoBEST as it offers a broad range of language understanding and reasoning tasks. KoBEST comprises five tasks: BoolQ, COPA, HellaSwag, WiC, and SentiNeg. However, given the findings of \citep{ko2023technical}, that both monolingual and multilingual language models exhibit inconsistent performance on WiC, we omit this task from our assessment. While other available datasets may be adopted as baselines, they come with limitations. For instance, Korean-NLI\&STS~\citep{ham-etal-2020-kornli}, being translated from English, is inherently more accessible for English models. KLUE~\citep{park2021klue}, despite being handcrafted, primarily focuses on basic NLU tasks like topic classification and NER. This makes it incapable of evaluating complex reasoning capabilities. Additionally, its test set is not publicly available.

\section{Evaluation Results}

\paragraph{Is HAE-RAE bench harder for foreign models?}~In Tables~\ref{evresults1} and \ref{evresults2}, we observe that the performance of LLMs scales with model size and the number of exemplars within the same suite. Nevertheless, despite their extensive training budgets, UMT5 and Llama-2 consistently fall short of their Polyglot-Ko counterparts. Furthermore, they rarely surpass the results of Polyglot-Ko-1.3B (0-shot). These results reaffirm the importance of language-specific corpora in learning cultural context and knowledge. It also highlights the effectiveness of HAE-RAE Bench, in assessing the language model's proficiency in Korean.

\begin{table}[htbp]
\centering
\fontsize{8}{10}\selectfont
\begin{tabular}{lcccc}
\toprule
& \multicolumn{2}{c}{\textbf{Polyglot-Ko}} & \multicolumn{2}{c}{\textbf{$\Delta$}}  \\
Dataset & Params & Average & UMT5 & Llama-2  \\ \midrule
\multirow{4}{*}{\begin{tabular}[c]{@{}l@{}}HAE-RAE\\ Bench\end{tabular}}   & 1.3B & 51.0 & -16.5 & -15.1  \\
 & 3.8B & 54.6 & -20.1 & -18.7 \\
 & 5.8B & 59.4 & -25.0 & -23.5  \\
 & 12.8B & 59.5 & -25.1 & -23.6  \\  
 \midrule
\multirow{4}{*}{KoBEST} & 1.3B & 56.3 & -6.3 & -6.1  \\
 & 3.8B & 55.7 & -5.7 & -5.5  \\
 &  5.8B & 56.0 & -6.0 & -5.8  \\
 & 12.8B & 65.2 & -15.2 & -15.0 \\ \bottomrule
\end{tabular}%
\caption{\footnotesize Average Performance of Polyglot-Ko vs. UMT5-XXL and Llama-2-13B on HAE-RAE and KoBEST (0-shot).}
\label{delta0}
\end{table}

\begin{table}[htbp]
\centering
\fontsize{8}{10}\selectfont
\begin{tabular}{lcccc}
\toprule
 & \multicolumn{2}{c}{\textbf{Polyglot-Ko}} & \multicolumn{2}{c}{$\Delta$}  \\
Dataset & Params & Average & UMT5 & Llama-2  \\ \midrule
\multirow{4}{*}{\begin{tabular}[c]{@{}l@{}}HAE-RAE\\ Bench\end{tabular}} & 1.3B & 61.1 & -27.4 & -13.7  \\
 & 3.8B & 66.8 & -33.1 & -19.4  \\
 & 5.8B & 67.4 & -33.7 & -20.0  \\
 & 12.8B & 70.4 & -36.6 & -22.9  \\ \midrule
\multirow{4}{*}{KoBEST} & 1.3B & 56.4 & -8.4 & {\color[HTML]{009901} 7.8} \\
 & 3.8B & 64.7 & -16.6 & -0.5  \\
 & 5.8B & 68.0 & -19.9 & -3.8  \\
 & 12.8B & 71.4 & -23.3 & -7.2  \\ \bottomrule
\end{tabular}%
\caption{Average Performance of Polyglot-Ko vs. UMT5-XXL and Llama-2-13B on HAE-RAE and KoBEST (5-shot).}
\label{delta5}
\end{table}

\begin{table}[htbp]
\centering
\fontsize{8}{10}\selectfont
\begin{tabular}{lcccc}
\toprule
 & \multicolumn{2}{c}{\textbf{Polyglot-Ko}} & \multicolumn{2}{c}{$\Delta$}  \\
Dataset & Params & Average & UMT5 & Llama-2 \\ \midrule
 \multirow{4}{*}{\begin{tabular}[c]{@{}l@{}}HAE-RAE\\ Bench\end{tabular}} & 1.3B & 61.2 & -28.0 & -12.9  \\
  & 3.8B & 66.7 & -33.4 & -18.4 \\
  & 5.8B & 67.3 & -34.1 & -19.0  \\
  & 12.8B & 71.6 & -38.3 & -23.2 \\ \midrule
\multirow{4}{*}{KoBEST} & 1.3B & 55.0 & -8.7 & {\color[HTML]{009901} 12.1}  \\
  & 3.8B & 63.3 & -16.9 & {\color[HTML]{009901} 3.9} \\
  & 5.8B & 68.0 & -21.6 & -0.8 \\
  & 12.8B & 71.8 & -25.4 & -4.6 \\ \bottomrule
\end{tabular}%
\caption{\footnotesize Average Performance of Polyglot-Ko vs. UMT5-XXL and Llama-2-13B on HAE-RAE and KoBEST (10-shot).}
\label{delta10}
\end{table}

Our results illustrated in Tables~\ref{delta0},~\ref{delta5}, and ~\ref{delta10} suggest that the HAE-RAE Bench is particularly challenging for non-Korean models compared to the KoBEST benchmark. The performance gap between Polyglot-Ko and its counterparts is more pronounced on the HAE-RAE Bench than KoBEST across all exemplar counts. Notably, for Llama-2-13B, the margin narrows considerably on KoBEST with an increase in exemplars. This discrepancy highlights that the proposed benchmark is especially challenging for models not tailored in Korean and difficult to mitigate by in-context learning. The entire result for KoBEST is illustrated in section~\ref{a:kobest}. \\

\paragraph{Does language frequency in the training corpora matter?} Despite UMT5 being trained on a larger volume of Korean tokens, it underperforms Llama-2 on the HAE-RAE Bench. Moreover, the advantage of in-context learning is relatively minimal for UMT5. Our findings support previous claims that language-specific reasoning capabilities of language models are not solely tied to the number of dedicated tokens in the pretraining corpus~\citep{shi2022language}. These results indicate that language models under the size of 20B parameters also transfer their in-context learning abilities to low-resource languages. \\ 

\paragraph{How important is the model size for the HAE-RAE Bench?} In Table~\ref{reganova}, we employ regression and Analysis of Variance(ANOVA) to examine the correlation between the parameter count of Polyglot-Ko models and their performance. To narrow the focus solely on the impact of model size, the analysis is limited to the Polyglot-Ko family, thus setting aside variables like corpus quality or model architecture. For the KoBEST benchmark, the results demonstrate a marked relationship between performance and model size, as indicated by the high $R^2$ value of 0.71 and the significant F-statistic and p-value. In contrast, for the HAE-RAE Bench, the model size explains only about a quarter of the performance variability. Additionally, the absence of statistical significance in both the regression and ANOVA for the HAE-RAE Bench implies that its evaluation is influenced by a broader spectrum of factors, pointing to challenges beyond just model size.

\begin{table}[htbp]
\centering
\fontsize{8}{11}\selectfont
\begin{tabular}{lcccc} \toprule
 & \multicolumn{3}{c}{\textbf{Regression}} & \textbf{ANOVA} \\ 
Benchmark & \( \beta_0 \) & \( \beta_1 \) & \( R^2 \) & \( F \)-statistic \\ \midrule
HAE-RAE Bench & 58.79 & 0.73 & 0.26 & 1.42 \\
KoBEST & 56.49 & 1.17 & 0.71* & 8.23* \\ \bottomrule
\end{tabular}%
\caption{\footnotesize Results from regression and ANOVA for the HAE-RAE and KoBEST benchmarks. An asterisk (*) denotes outcomes with a p-value less than 0.01, indicating statistical significance.}
\label{reganova}
\end{table}

\paragraph{Can GPT-3.5/4 ace HAE-RAE Bench?}~In Table~\ref{gpt-ev}, the performance of GPT-3.5 and GPT-4 on the HAE-RAE Bench and KoBEST is presented. Unlike openly available models for which we leveraged a log probability method to gauge accuracy, these models do not provide log probabilities for individual tokens. Accordingly, we prompted the models to generate the number of the options they deemed correct. Direct comparison between these evaluation methods is not feasible. However, the method used for proprietary models is more challenging than the log-likelihood method applied to open models. The former entails generating answers from the entire vocabulary, whereas the latter restricts choices to five options. Notably, GPT-3.5 and GPT-4 achieved scores of 51.2\% and 67.8\% on the HAE-RAE Bench, respectively, indicating potential for further improvements. Conversely, their performances on KoBEST were 68.0\% and 81.1\%, suggesting narrower margins for improvement. In summary, state-of-the-art language models such as GPT-3.5 and GPT-4 have yet to master either the HAE-RAE Bench or KoBEST, though more room is left for the HAE-RAE Bench. The entire evaluation results for GPT-3.5 and GPT-4 models is available at section~\ref{a:gpt}.

\begin{table*}[htbp]
\centering
\fontsize{9}{11}\selectfont
\begin{tabular}{lcccccc}
\toprule
 & \multicolumn{3}{c}{\textbf{GPT-3.5-Turbo}} &  \multicolumn{3}{c}{\textbf{GPT-4}} \\ \cmidrule(lr){2-4} \cmidrule(lr){5-7}
Dataset & Ko & En & $\Delta$ & Ko & En & $\Delta$ \\ \midrule
HAE-RAE Bench & 51.2 & 55.4 & {\color[HTML]{009901} 4.2} &  67.8 & 68.2 & {\color[HTML]{009901} 0.4} \\
KoBEST & 68.0 & 79.3 & {\color[HTML]{009901} 11.4} &  81.1 & 91.0 & {\color[HTML]{009901} 9.9} \\ \bottomrule
\end{tabular}%
\caption{\footnotesize Evaluation result of GPT-3.5-Turbo and GPT-4 on HAE-RAE Bench and KoBEST with zero shot setting. We use the snapshot from June 13th 2023 for both models. Ko and En denote the language of the prompt used.}
\label{gpt-ev}
\end{table*}

\paragraph{Can knowledge be transferred from English?}~Past research indicates that LLMs can internally transfer knowledge acquired in English to low-resource languages~\citep{ huang2023not,zhou2023enhancing}. We employ Cross-lingual thought prompting (XLT) with GPT-3.5-Turbo and GPT-4 to investigate whether LLMs leverage abilities derived from English corpora to solve HAE-RA Bench. XLT~\citep{shi2022language} is a technique that aids the transfer of abilities learned in English to other languages. As illustrated in Table~\ref{gpt-ev}, English prompting enhances the performance of LLMs on both the HAE-RAE Bench and KoBEST. However, the gains for the HAE-RAE Bench are modest: 4.2 for GPT-3.5-Turbo and 0.4 for GPT-4. In contrast, the improvements on KoBEST are more substantial, with margins of 9.9 and 11.1, respectively. Given KoBEST's focus on language understanding and reasoning, we suspect that such skills are more seamlessly transferable across languages within models. On the other hand, the HAE-RAE Bench probes the nuances of cultural context and knowledge, which are challenging to learn from English tokens, thereby undermining the benefits of extensive training across various languages.

\section{Error Analysis}

Error analysis is essential to understand the common errors or likely biases of language model mistakes and identify areas of future research. Accordingly, we compare the results of Polyglot-Ko-12.8B (0-shot) and GPT-4 (Korean Prompting) for possible errors. 

\begin{figure}[t]
\includegraphics[width=\columnwidth]{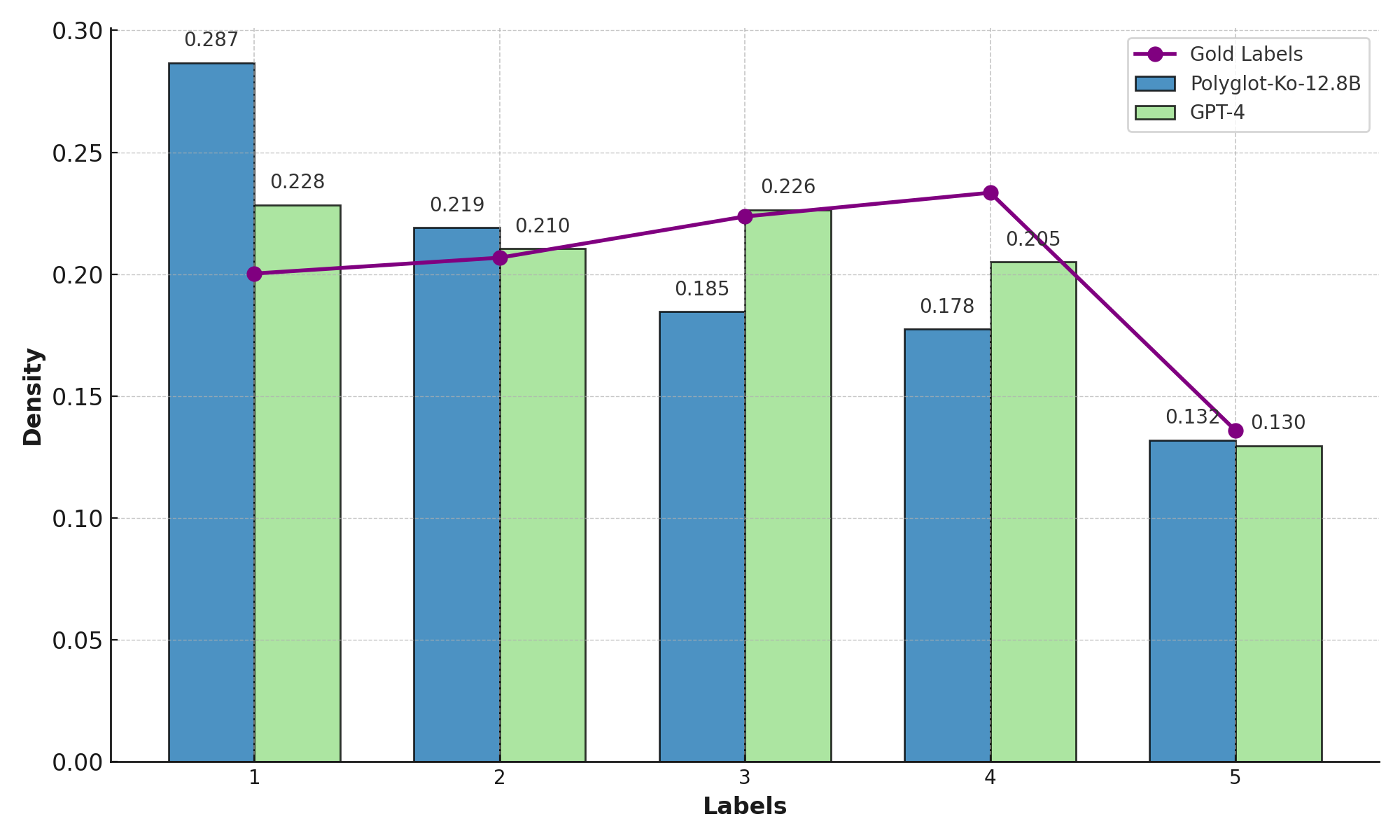}
\centering
\caption{\footnotesize Density distribution of answer choices by Polyglot-Ko-12.8B, GPT-4, and Gold Labels.}
\label{dist}
\end{figure}

We first examine the answer distribution to see if either model has a bias toward selecting certain numbers. This is shown in Figure~\ref{dist}. We find that both models are less likely to guess `` 5" compared to other numbers. This pattern can be traced back to the dataset composition: while most questions offer five multiple-choice options, the reading comprehension subset provides only four. Beyond this, neither model displays any notable trends.

\begin{table*}[htbp]
\centering
\fontsize{9}{11}\selectfont
\begin{tabular}{ccccc}
\toprule
 &  \multicolumn{2}{c}{GPT-4} & \multicolumn{2}{c}{Polyglot-Ko-12.8B} \\ 
Dataset &  \multicolumn{1}{c}{Correct} & \multicolumn{1}{c}{Incorrect} &  \multicolumn{1}{c}{Correct} & \multicolumn{1}{c}{Incorrect} \\ \midrule
Rare Words &  1.58 (0.81) & 1.58 (0.80) &  1.58 (0.81) & 1.58 (0.80) \\
Loan Words &  1.58 (0.80) & 1.56 (0.80) &  1.58 (0.80) & 1.57 (0.80) \\
Standard Nomenclature &  1.58 (0.80) & 1.55 (0.80) &  1.57 (0.81) & 1.56 (0.80) \\ \bottomrule
\end{tabular}%
\caption{\footnotesize Average Levenshtein distance of options for correct and incorrect questions.}
\label{vocabl}
\end{table*}

In the \textbf{Rare Words}, \textbf{Loan Words}, and \textbf{Standard Nomenclature} subsets of the HAE-RAE Bench, incorrect options were generated using a sorting method based on Levenshtein distance. To investigate the impact of Levenshtein distance on model performance, we compare the average distances for options based on whether the model answered the question correctly. As shown in Table~\ref{vocabl}, no discernible difference in Levenshtein distance is observed for either model between correct and incorrect answers. We suspect the set Levenshtein distance threshold of 3 may not lead to meaningful variations in question difficulty. We review all incorrect questions for Polyglot-Ko-12.8B and GPT-4 to delve deeper. However, given the questions' simple structures, such as \textit{"What is the \{official loan word / correct standard nomenclature\} for \{word\}?"} or \textit{"Which word is suitable for the definition \{def\}?"}, we did not identify any syntactic characteristics that might explain the incorrect questions.

\begin{figure}[t]
\includegraphics[width=\columnwidth]{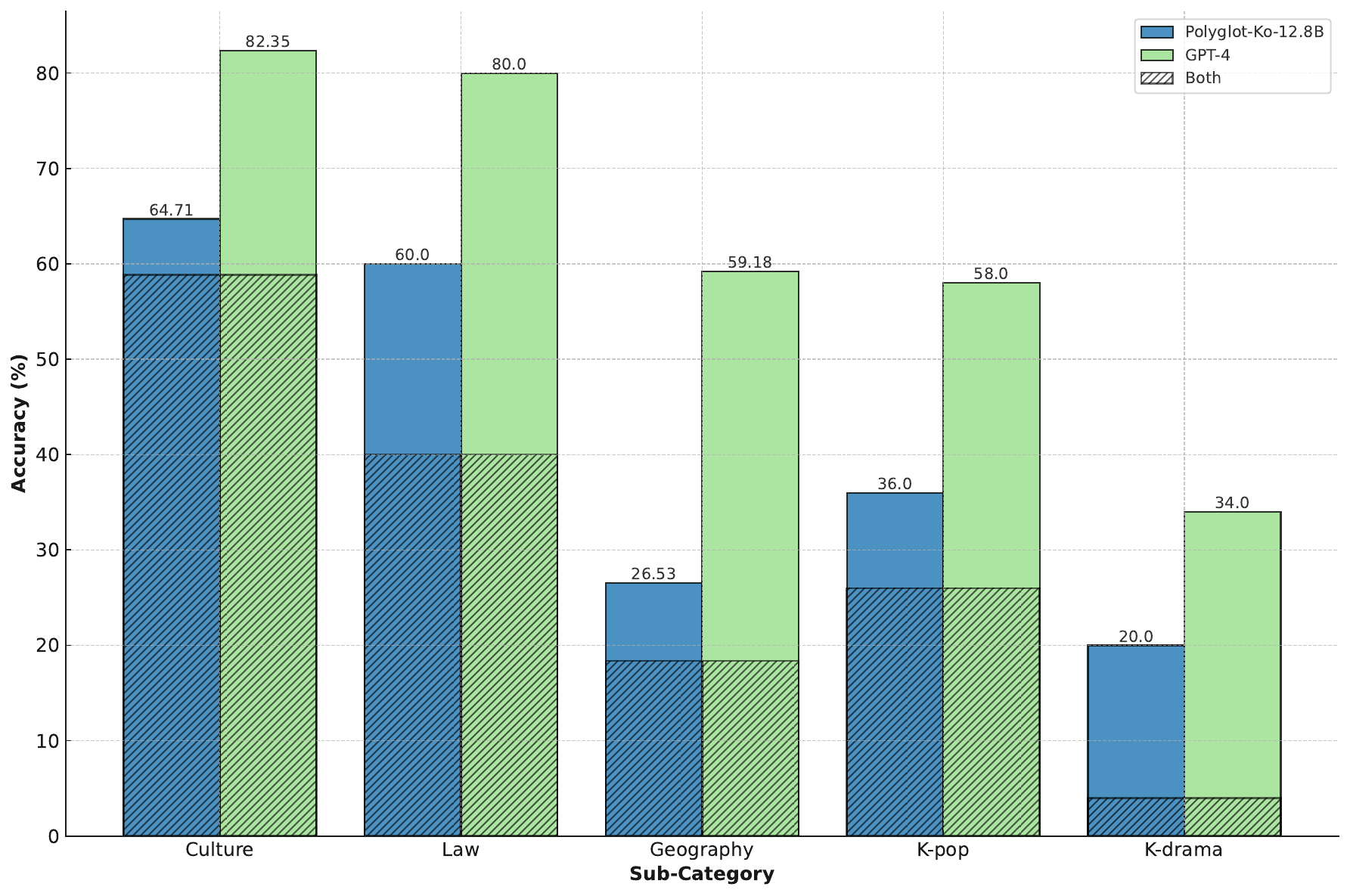}
\centering
\caption{\footnotesize Accuracy of  Polyglot-Ko-12.8B and GPT-4 on sub-categories of General Knowledge. The striped areas within each bar represent questions that both models answered correctly.}
\label{kgkd}
\end{figure}

\begin{figure}[htbp]
\includegraphics[width=\columnwidth]{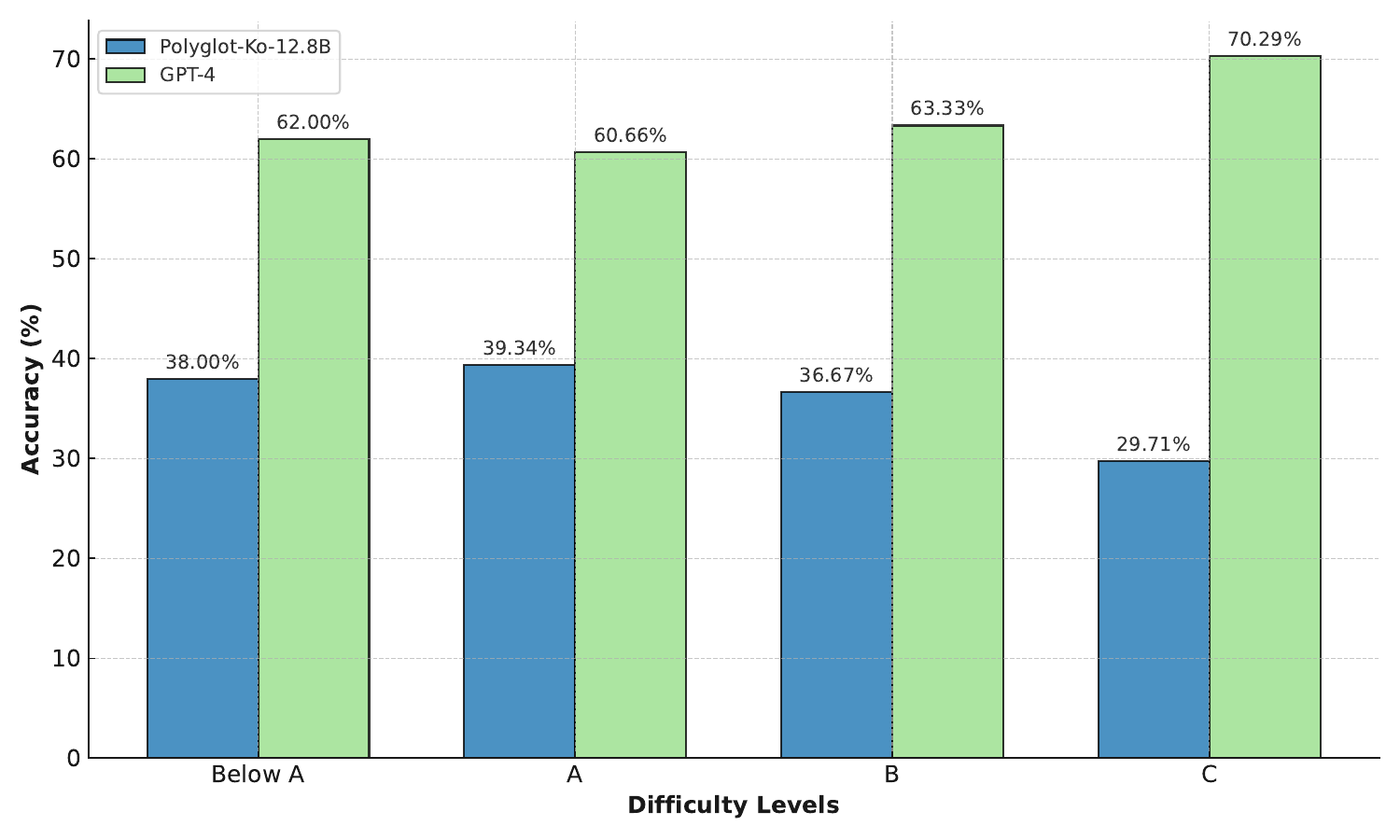}
\centering
\caption{\footnotesize Accuracy of Polyglot-Ko-12.8B and GPT-4 on different levels of Reading Comprehension.}
\label{rcd}
\end{figure}

For the \textbf{General Knowledge} subset, we assess the performance of Polyglot-Ko-12.8B and GPT-4 across the subcategories, as shown in Figure~\ref{kgkd}. GPT-4 consistently outperforms Polyglot-Ko-12.8B. Polyglot-Ko-12.8B fares better in law and culture but lags in geography, K-pop, and especially K-drama. Given the need for up-to-date information in these areas, the model's weaker performance in K-pop and K-drama may stem from its knowledge cutoff. GPT-4 excels across all categories, likely benefiting from a diverse training set. Both models have the lowest scores in the K-drama category, suggesting either model limitations or ambiguous questions. 

The \textbf{Reading Comprehension} subset of HAE-RAE Bench comprises four difficulty levels: Introductory (for absolute beginners), A (beginner), B (intermediate), and C (advanced), based on the Common European Framework of Reference (CEFR). In figure~\ref{rcd}, we examine the performance of each model across these levels. Our findings indicate that GPT-4 consistently outperforms Polyglot-Ko-12.8B across all difficulty tiers, with the performance gap becoming more pronounced at higher levels (B and C). The performance of Polyglot-Ko-12.8B peaks at difficulty level A and declines, suggesting a limitation in handling more challenging questions.

\section{License}

HAE-RAE Bench is released under a CC BY-NC-ND license. This license prohibits remixing, redistribution, and commercial use of the dataset. This constraint is due to the reading comprehension subset, for which the copyright holder of KLAT has restricted commercial alterations. However, we do not anticipate this as a significant issue since benchmark datasets are rarely used for commercial purposes. Researchers can still freely download and evaluate their models using this dataset.

\section{Conclusion}

This paper introduces the HAE-RAE Bench, a dataset curated to evaluate the cultural knowledge encoded in language models. Unlike previous Korean language model evaluation suites, the HAE-RAE Bench is crafted to present a greater challenge to non-Korean models, disrupting their ability to guess answers based on in-context learning or scale-derived multilingualism. Our work is among the first to propose a non-task-oriented dataset aimed at assessing whether a language model's knowledge is adequate for roles like a domestic conversational agent or search engine. This research suggests a pathway for advancing non-English NLP, emphasizing the need for language models that are as proficient in language-specific knowledge as they are in language understanding and reasoning tasks.

\section*{Acknowledgments}
This research was supported by Brian Impact Foundation, a non-profit organization dedicated to the advancement of science and technology for all.

\nocite{*}
\section{Bibliographical References}\label{sec:reference}

\bibliographystyle{lrec-coling2024-natbib}
\bibliography{HRB}

\begin{thebibliography}{52}
\expandafter\ifx\csname natexlab\endcsname\relax\def\natexlab#1{#1}\fi

\bibitem[{ai~forever(2023)}]{rugpt}
ai~forever. 2023.
\newblock \href {https://huggingface.co/ai-forever/ruGPT-3.5-13B} {rugpt-3.5
  13b}.

\bibitem[{Akhbardeh et~al.(2021)Akhbardeh, Arkhangorodsky, Biesialska, Bojar,
  Chatterjee, Chaudhary, Costa-jussa, Espa{\~n}a-Bonet, Fan, Federmann,
  Freitag, Graham, Grundkiewicz, Haddow, Harter, Heafield, Homan, Huck,
  Amponsah-Kaakyire, Kasai, Khashabi, Knight, Kocmi, Koehn, Lourie, Monz,
  Morishita, Nagata, Nagesh, Nakazawa, Negri, Pal, Tapo, Turchi, Vydrin, and
  Zampieri}]{akhbardeh-etal-2021-findings}
Farhad Akhbardeh, Arkady Arkhangorodsky, Magdalena Biesialska, Ond{\v{r}}ej
  Bojar, Rajen Chatterjee, Vishrav Chaudhary, Marta~R. Costa-jussa, Cristina
  Espa{\~n}a-Bonet, Angela Fan, Christian Federmann, Markus Freitag, Yvette
  Graham, Roman Grundkiewicz, Barry Haddow, Leonie Harter, Kenneth Heafield,
  Christopher Homan, Matthias Huck, Kwabena Amponsah-Kaakyire, Jungo Kasai,
  Daniel Khashabi, Kevin Knight, Tom Kocmi, Philipp Koehn, Nicholas Lourie,
  Christof Monz, Makoto Morishita, Masaaki Nagata, Ajay Nagesh, Toshiaki
  Nakazawa, Matteo Negri, Santanu Pal, Allahsera~Auguste Tapo, Marco Turchi,
  Valentin Vydrin, and Marcos Zampieri. 2021.
\newblock \href {https://aclanthology.org/2021.wmt-1.1} {Findings of the 2021
  conference on machine translation ({WMT}21)}.
\newblock In \emph{Proceedings of the Sixth Conference on Machine Translation},
  pages 1--88, Online. Association for Computational Linguistics.

\bibitem[{Anil et~al.(2023)Anil, Dai, Firat, Johnson, Lepikhin, Passos,
  Shakeri, Taropa, Bailey, Chen et~al.}]{anil2023palm}
Rohan Anil, Andrew~M Dai, Orhan Firat, Melvin Johnson, Dmitry Lepikhin,
  Alexandre Passos, Siamak Shakeri, Emanuel Taropa, Paige Bailey, Zhifeng Chen,
  et~al. 2023.
\newblock Palm 2 technical report.
\newblock \emph{arXiv preprint arXiv:2305.10403}.

\bibitem[{Chung et~al.(2023)Chung, Constant, Garcia, Roberts, Tay, Narang, and
  Firat}]{chung2023unimax}
Hyung~Won Chung, Noah Constant, Xavier Garcia, Adam Roberts, Yi~Tay, Sharan
  Narang, and Orhan Firat. 2023.
\newblock Unimax: Fairer and more effective language sampling for large-scale
  multilingual pretraining.
\newblock \emph{arXiv preprint arXiv:2304.09151}.

\bibitem[{Chung et~al.(2022)Chung, Hou, Longpre, Zoph, Tay, Fedus, Li, Wang,
  Dehghani, Brahma et~al.}]{flant5}
Hyung~Won Chung, Le~Hou, Shayne Longpre, Barret Zoph, Yi~Tay, William Fedus,
  Eric Li, Xuezhi Wang, Mostafa Dehghani, Siddhartha Brahma, et~al. 2022.
\newblock Scaling instruction-finetuned language models.
\newblock \emph{arXiv preprint arXiv:2210.11416}.

\bibitem[{Clark et~al.(2019)Clark, Lee, Chang, Kwiatkowski, Collins, and
  Toutanova}]{clark-etal-2019-boolq}
Christopher Clark, Kenton Lee, Ming-Wei Chang, Tom Kwiatkowski, Michael
  Collins, and Kristina Toutanova. 2019.
\newblock \href {https://doi.org/10.18653/v1/N19-1300} {{B}ool{Q}: Exploring
  the surprising difficulty of natural yes/no questions}.
\newblock In \emph{Proceedings of the 2019 Conference of the North {A}merican
  Chapter of the Association for Computational Linguistics: Human Language
  Technologies, Volume 1 (Long and Short Papers)}, pages 2924--2936,
  Minneapolis, Minnesota. Association for Computational Linguistics.

\bibitem[{Clark et~al.(2020)Clark, Choi, Collins, Garrette, Kwiatkowski,
  Nikolaev, and Palomaki}]{clark-etal-2020-tydi}
Jonathan~H. Clark, Eunsol Choi, Michael Collins, Dan Garrette, Tom Kwiatkowski,
  Vitaly Nikolaev, and Jennimaria Palomaki. 2020.
\newblock \href {https://doi.org/10.1162/tacl_a_00317} {{T}y{D}i {QA}: A
  benchmark for information-seeking question answering in typologically diverse
  languages}.
\newblock \emph{Transactions of the Association for Computational Linguistics},
  8:454--470.

\bibitem[{Conneau et~al.(2018)Conneau, Lample, Rinott, Williams, Bowman,
  Schwenk, and Stoyanov}]{conneau2018xnli}
Alexis Conneau, Guillaume Lample, Ruty Rinott, Adina Williams, Samuel~R Bowman,
  Holger Schwenk, and Veselin Stoyanov. 2018.
\newblock Xnli: Evaluating cross-lingual sentence representations.
\newblock \emph{arXiv preprint arXiv:1809.05053}.

\bibitem[{Cui et~al.(2023)Cui, Yang, and Yao}]{cui2023efficient}
Yiming Cui, Ziqing Yang, and Xin Yao. 2023.
\newblock Efficient and effective text encoding for chinese llama and alpaca.
\newblock \emph{arXiv preprint arXiv:2304.08177}.

\bibitem[{Devlin et~al.(2018)Devlin, Chang, Lee, and
  Toutanova}]{devlin2018bert}
Jacob Devlin, Ming-Wei Chang, Kenton Lee, and Kristina Toutanova. 2018.
\newblock Bert: Pre-training of deep bidirectional transformers for language
  understanding.
\newblock \emph{arXiv preprint arXiv:1810.04805}.

\bibitem[{Gao et~al.(2021)Gao, Tow, Biderman, Black, DiPofi, Foster, Golding,
  Hsu, McDonell, Muennighoff, Phang, Reynolds, Tang, Thite, Wang, Wang, and
  Zou}]{eval-harness}
Leo Gao, Jonathan Tow, Stella Biderman, Sid Black, Anthony DiPofi, Charles
  Foster, Laurence Golding, Jeffrey Hsu, Kyle McDonell, Niklas Muennighoff,
  Jason Phang, Laria Reynolds, Eric Tang, Anish Thite, Ben Wang, Kevin Wang,
  and Andy Zou. 2021.
\newblock \href {https://doi.org/10.5281/zenodo.5371628} {A framework for
  few-shot language model evaluation}.

\bibitem[{Gordon et~al.(2012)Gordon, Kozareva, and
  Roemmele}]{gordon-etal-2012-semeval}
Andrew Gordon, Zornitsa Kozareva, and Melissa Roemmele. 2012.
\newblock \href {https://aclanthology.org/S12-1052} {{S}em{E}val-2012 task 7:
  Choice of plausible alternatives: An evaluation of commonsense causal
  reasoning}.
\newblock In \emph{*{SEM} 2012: The First Joint Conference on Lexical and
  Computational Semantics {--} Volume 1: Proceedings of the main conference and
  the shared task, and Volume 2: Proceedings of the Sixth International
  Workshop on Semantic Evaluation ({S}em{E}val 2012)}, pages 394--398,
  Montr{\'e}al, Canada. Association for Computational Linguistics.

\bibitem[{Ham et~al.(2020)Ham, Choe, Park, Choi, and
  Soh}]{ham-etal-2020-kornli}
Jiyeon Ham, Yo~Joong Choe, Kyubyong Park, Ilji Choi, and Hyungjoon Soh. 2020.
\newblock \href {https://doi.org/10.18653/v1/2020.findings-emnlp.39}
  {{K}or{NLI} and {K}or{STS}: New benchmark datasets for {K}orean natural
  language understanding}.
\newblock In \emph{Findings of the Association for Computational Linguistics:
  EMNLP 2020}, pages 422--430, Online. Association for Computational
  Linguistics.

\bibitem[{Hendrycks et~al.(2020)Hendrycks, Burns, Basart, Zou, Mazeika, Song,
  and Steinhardt}]{hendrycks2020mmlu}
Dan Hendrycks, Collin Burns, Steven Basart, Andy Zou, Mantas Mazeika, Dawn
  Song, and Jacob Steinhardt. 2020.
\newblock Measuring massive multitask language understanding.
\newblock \emph{arXiv preprint arXiv:2009.03300}.

\bibitem[{Huang et~al.(2023)Huang, Tang, Zhang, Zhao, Song, Xia, and
  Wei}]{huang2023not}
Haoyang Huang, Tianyi Tang, Dongdong Zhang, Wayne~Xin Zhao, Ting Song, Yan Xia,
  and Furu Wei. 2023.
\newblock Not all languages are created equal in llms: Improving multilingual
  capability by cross-lingual-thought prompting.
\newblock \emph{arXiv preprint arXiv:2305.07004}.

\bibitem[{Kaushik and Lipton(2018)}]{kaushik2018much}
Divyansh Kaushik and Zachary~C Lipton. 2018.
\newblock How much reading does reading comprehension require? a critical
  investigation of popular benchmarks.
\newblock \emph{arXiv preprint arXiv:1808.04926}.

\bibitem[{Kim et~al.(2021)Kim, Kim, Lee, Lee, Kwak, Dong~Hyeon, Park, Kim, Kim,
  Seo, Lee, Jeong, Lee, Kim, Ko, Kim, Park, Kim, Kang, Ryu, Yoo, Chang, Suh,
  In, Park, Kim, Kim, Jeong, Yeo, Ham, Park, Lee, Kang, Kang, Ha, Park, and
  Sung}]{kim-etal-2021-changes}
Boseop Kim, HyoungSeok Kim, Sang-Woo Lee, Gichang Lee, Donghyun Kwak, Jeon
  Dong~Hyeon, Sunghyun Park, Sungju Kim, Seonhoon Kim, Dongpil Seo, Heungsub
  Lee, Minyoung Jeong, Sungjae Lee, Minsub Kim, Suk~Hyun Ko, Seokhun Kim,
  Taeyong Park, Jinuk Kim, Soyoung Kang, Na-Hyeon Ryu, Kang~Min Yoo, Minsuk
  Chang, Soobin Suh, Sookyo In, Jinseong Park, Kyungduk Kim, Hiun Kim, Jisu
  Jeong, Yong~Goo Yeo, Donghoon Ham, Dongju Park, Min~Young Lee, Jaewook Kang,
  Inho Kang, Jung-Woo Ha, Woomyoung Park, and Nako Sung. 2021.
\newblock \href {https://doi.org/10.18653/v1/2021.emnlp-main.274} {What changes
  can large-scale language models bring? intensive study on {H}yper{CLOVA}:
  Billions-scale {K}orean generative pretrained transformers}.
\newblock In \emph{Proceedings of the 2021 Conference on Empirical Methods in
  Natural Language Processing}, pages 3405--3424, Online and Punta Cana,
  Dominican Republic. Association for Computational Linguistics.

\bibitem[{Kim et~al.(2022)Kim, Jang, Kwon, and Davis}]{kim2022kobest}
Dohyeong Kim, Myeongjun Jang, Deuk~Sin Kwon, and Eric Davis. 2022.
\newblock Kobest: Korean balanced evaluation of significant tasks.
\newblock \emph{arXiv preprint arXiv:2204.04541}.

\bibitem[{Ko et~al.(2023)Ko, Yang, Ryu, Choi, Yang, Park
  et~al.}]{ko2023technical}
Hyunwoong Ko, Kichang Yang, Minho Ryu, Taekyoon Choi, Seungmu Yang, Sungho
  Park, et~al. 2023.
\newblock A technical report for polyglot-ko: Open-source large-scale korean
  language models.
\newblock \emph{arXiv preprint arXiv:2306.02254}.

\bibitem[{Kurihara et~al.(2022)Kurihara, Kawahara, and
  Shibata}]{kurihara-etal-2022-jglue}
Kentaro Kurihara, Daisuke Kawahara, and Tomohide Shibata. 2022.
\newblock \href {https://aclanthology.org/2022.lrec-1.317} {{JGLUE}: {J}apanese
  general language understanding evaluation}.
\newblock In \emph{Proceedings of the Thirteenth Language Resources and
  Evaluation Conference}, pages 2957--2966, Marseille, France. European
  Language Resources Association.

\bibitem[{Lee et~al.(2023)Lee, Hunter, and Ruiz}]{lee2023platypus}
Ariel~N Lee, Cole~J Hunter, and Nataniel Ruiz. 2023.
\newblock Platypus: Quick, cheap, and powerful refinement of llms.
\newblock \emph{arXiv preprint arXiv:2308.07317}.

\bibitem[{Leong et~al.(2023)Leong, Ngui, Susanto, Rengarajan, Sarveswaran, and
  Tjhi}]{leong2023bhasa}
Wei~Qi Leong, Jian~Gang Ngui, Yosephine Susanto, Hamsawardhini Rengarajan,
  Kengatharaiyer Sarveswaran, and William~Chandra Tjhi. 2023.
\newblock \href {http://arxiv.org/abs/2309.06085} {Bhasa: A holistic southeast
  asian linguistic and cultural evaluation suite for large language models}.

\bibitem[{Levenshtein et~al.(1966)}]{levenshtein1966binary}
Vladimir~I Levenshtein et~al. 1966.
\newblock Binary codes capable of correcting deletions, insertions, and
  reversals.
\newblock In \emph{Soviet physics doklady}, volume~10, pages 707--710. Soviet
  Union.

\bibitem[{Li et~al.(2023)Li, Zhang, Koto, Yang, Zhao, Gong, Duan, and
  Baldwin}]{li2023cmmlu}
Haonan Li, Yixuan Zhang, Fajri Koto, Yifei Yang, Hai Zhao, Yeyun Gong, Nan
  Duan, and Timothy Baldwin. 2023.
\newblock Cmmlu: Measuring massive multitask language understanding in chinese.
\newblock \emph{arXiv preprint arXiv:2306.09212}.

\bibitem[{Liang et~al.(2022)Liang, Bommasani, Lee, Tsipras, Soylu, Yasunaga,
  Zhang, Narayanan, Wu, Kumar et~al.}]{liang2022helm}
Percy Liang, Rishi Bommasani, Tony Lee, Dimitris Tsipras, Dilara Soylu,
  Michihiro Yasunaga, Yian Zhang, Deepak Narayanan, Yuhuai Wu, Ananya Kumar,
  et~al. 2022.
\newblock Holistic evaluation of language models.
\newblock \emph{arXiv preprint arXiv:2211.09110}.

\bibitem[{Malaviya et~al.(2022)Malaviya, Bhatia, and
  Yatskar}]{malaviya-etal-2022-cascading}
Chaitanya Malaviya, Sudeep Bhatia, and Mark Yatskar. 2022.
\newblock \href {https://doi.org/10.18653/v1/2022.emnlp-main.438} {Cascading
  biases: Investigating the effect of heuristic annotation strategies on data
  and models}.
\newblock In \emph{Proceedings of the 2022 Conference on Empirical Methods in
  Natural Language Processing}, pages 6525--6540, Abu Dhabi, United Arab
  Emirates. Association for Computational Linguistics.

\bibitem[{OpenAI(2023)}]{OpenAI2023GPT4TR}
OpenAI. 2023.
\newblock \href {https://api.semanticscholar.org/CorpusID:257532815} {Gpt-4
  technical report}.
\newblock \emph{ArXiv}, abs/2303.08774.

\bibitem[{Ouyang et~al.(2022)Ouyang, Wu, Jiang, Almeida, Wainwright, Mishkin,
  Zhang, Agarwal, Slama, Ray et~al.}]{instructgpt}
Long Ouyang, Jeffrey Wu, Xu~Jiang, Diogo Almeida, Carroll Wainwright, Pamela
  Mishkin, Chong Zhang, Sandhini Agarwal, Katarina Slama, Alex Ray, et~al.
  2022.
\newblock Training language models to follow instructions with human feedback.
\newblock \emph{Advances in Neural Information Processing Systems},
  35:27730--27744.

\bibitem[{Park et~al.(2021)Park, Moon, Kim, Cho, Han, Park, Song, Kim, Song, Oh
  et~al.}]{park2021klue}
Sungjoon Park, Jihyung Moon, Sungdong Kim, Won~Ik Cho, Jiyoon Han, Jangwon
  Park, Chisung Song, Junseong Kim, Yongsook Song, Taehwan Oh, et~al. 2021.
\newblock Klue: Korean language understanding evaluation.
\newblock \emph{arXiv preprint arXiv:2105.09680}.

\bibitem[{Penedo et~al.(2023)Penedo, Malartic, Hesslow, Cojocaru, Cappelli,
  Alobeidli, Pannier, Almazrouei, and Launay}]{penedo2023refinedweb}
Guilherme Penedo, Quentin Malartic, Daniel Hesslow, Ruxandra Cojocaru,
  Alessandro Cappelli, Hamza Alobeidli, Baptiste Pannier, Ebtesam Almazrouei,
  and Julien Launay. 2023.
\newblock The refinedweb dataset for falcon llm: outperforming curated corpora
  with web data, and web data only.
\newblock \emph{arXiv preprint arXiv:2306.01116}.

\bibitem[{Pilehvar and Camacho-Collados(2018)}]{pilehvar2018wic}
Mohammad~Taher Pilehvar and Jose Camacho-Collados. 2018.
\newblock Wic: the word-in-context dataset for evaluating context-sensitive
  meaning representations.
\newblock \emph{arXiv preprint arXiv:1808.09121}.

\bibitem[{Pires et~al.(2023)Pires, Abonizio, Rog{\'e}rio, and
  Nogueira}]{pires2023sabi}
Ramon Pires, Hugo Abonizio, Thales Rog{\'e}rio, and Rodrigo Nogueira. 2023.
\newblock Sabi$\backslash$'a: Portuguese large language models.
\newblock \emph{arXiv preprint arXiv:2304.07880}.

\bibitem[{QwenLM(2023)}]{qwen}
QwenLM. 2023.
\newblock \href {https://github.com/QwenLM/Qwen-7B/blob/main/tech_memo.md}
  {Introducing qwen-7b: Open foundation and human-aligned models (of the
  state-of-the-arts)}.

\bibitem[{Radford et~al.()Radford, Narasimhan, Salimans, Sutskever
  et~al.}]{radford2018improving}
Alec Radford, Karthik Narasimhan, Tim Salimans, Ilya Sutskever, et~al.
\newblock Improving language understanding by generative pre-training.

\bibitem[{Savanur and Sumathi(2023)}]{sentineg}
Sandhya Savanur and R.~Sumathi. 2023.
\newblock \href {https://doi.org/10.4018/IJSI.315741} {Sentineg: Algorithm to
  process negations at sentence level in sentiment analysis}.
\newblock \emph{International Journal of Software Innovation}, 11:1--27.

\bibitem[{Scao et~al.(2022)Scao, Fan, Akiki, Pavlick, Ili{\'c}, Hesslow,
  Castagn{\'e}, Luccioni, Yvon, Gall{\'e} et~al.}]{scao2022bloom}
Teven~Le Scao, Angela Fan, Christopher Akiki, Ellie Pavlick, Suzana Ili{\'c},
  Daniel Hesslow, Roman Castagn{\'e}, Alexandra~Sasha Luccioni, Fran{\c{c}}ois
  Yvon, Matthias Gall{\'e}, et~al. 2022.
\newblock Bloom: A 176b-parameter open-access multilingual language model.
\newblock \emph{arXiv preprint arXiv:2211.05100}.

\bibitem[{Shi et~al.(2022)Shi, Suzgun, Freitag, Wang, Srivats, Vosoughi, Chung,
  Tay, Ruder, Zhou et~al.}]{shi2022language}
Freda Shi, Mirac Suzgun, Markus Freitag, Xuezhi Wang, Suraj Srivats, Soroush
  Vosoughi, Hyung~Won Chung, Yi~Tay, Sebastian Ruder, Denny Zhou, et~al. 2022.
\newblock Language models are multilingual chain-of-thought reasoners.
\newblock \emph{arXiv preprint arXiv:2210.03057}.

\bibitem[{Son et~al.(2023)Son, Lee, Kang, and Hahm}]{son2023removing}
Guijin Son, Hanwool Lee, Nahyeon Kang, and Moonjeong Hahm. 2023.
\newblock Removing non-stationary knowledge from pre-trained language models
  for entity-level sentiment classification in finance.
\newblock \emph{arXiv preprint arXiv:2301.03136}.

\bibitem[{Srivastava et~al.(2022)Srivastava, Rastogi, Rao, Shoeb, Abid, Fisch,
  Brown, Santoro, Gupta, Garriga-Alonso et~al.}]{srivastava2022big}
Aarohi Srivastava, Abhinav Rastogi, Abhishek Rao, Abu Awal~Md Shoeb, Abubakar
  Abid, Adam Fisch, Adam~R Brown, Adam Santoro, Aditya Gupta, Adri{\`a}
  Garriga-Alonso, et~al. 2022.
\newblock Beyond the imitation game: Quantifying and extrapolating the
  capabilities of language models.
\newblock \emph{arXiv preprint arXiv:2206.04615}.

\bibitem[{StabilityAI(2023)}]{jlm}
StabilityAI. 2023.
\newblock \href
  {https://huggingface.co/stabilityai/japanese-stablelm-base-alpha-7b}
  {Japanese-stablelm-base-alpha-7b}.

\bibitem[{Touvron et~al.(2023)Touvron, Martin, Stone, Albert, Almahairi,
  Babaei, Bashlykov, Batra, Bhargava, Bhosale et~al.}]{touvron2023llama}
Hugo Touvron, Louis Martin, Kevin Stone, Peter Albert, Amjad Almahairi, Yasmine
  Babaei, Nikolay Bashlykov, Soumya Batra, Prajjwal Bhargava, Shruti Bhosale,
  et~al. 2023.
\newblock Llama 2: Open foundation and fine-tuned chat models.
\newblock \emph{arXiv preprint arXiv:2307.09288}.

\bibitem[{Vaswani et~al.(2017)Vaswani, Shazeer, Parmar, Uszkoreit, Jones,
  Gomez, Kaiser, and Polosukhin}]{vaswani2017attention}
Ashish Vaswani, Noam Shazeer, Niki Parmar, Jakob Uszkoreit, Llion Jones,
  Aidan~N Gomez, {\L}ukasz Kaiser, and Illia Polosukhin. 2017.
\newblock Attention is all you need.
\newblock \emph{Advances in neural information processing systems}, 30.

\bibitem[{Vicuna(2023)}]{vicuna}
Vicuna. 2023.
\newblock \href {https://lmsys.org/blog/2023-03-30-vicuna/} {Vicuna: An
  open-source chatbot impressing gpt-4 with 90\% chatgpt quality}.

\bibitem[{Wang et~al.(2019)Wang, Pruksachatkun, Nangia, Singh, Michael, Hill,
  Levy, and Bowman}]{wang2019superglue}
Alex Wang, Yada Pruksachatkun, Nikita Nangia, Amanpreet Singh, Julian Michael,
  Felix Hill, Omer Levy, and Samuel Bowman. 2019.
\newblock Superglue: A stickier benchmark for general-purpose language
  understanding systems.
\newblock \emph{Advances in neural information processing systems}, 32.

\bibitem[{Wang et~al.(2018)Wang, Singh, Michael, Hill, Levy, and
  Bowman}]{wang2018glue}
Alex Wang, Amanpreet Singh, Julian Michael, Felix Hill, Omer Levy, and Samuel~R
  Bowman. 2018.
\newblock Glue: A multi-task benchmark and analysis platform for natural
  language understanding.
\newblock \emph{arXiv preprint arXiv:1804.07461}.

\bibitem[{Xu et~al.(2023)Xu, Sun, Zheng, Geng, Zhao, Feng, Tao, and
  Jiang}]{xu2023wizardlm}
Can Xu, Qingfeng Sun, Kai Zheng, Xiubo Geng, Pu~Zhao, Jiazhan Feng, Chongyang
  Tao, and Daxin Jiang. 2023.
\newblock Wizardlm: Empowering large language models to follow complex
  instructions.
\newblock \emph{arXiv preprint arXiv:2304.12244}.

\bibitem[{Xue et~al.(2020)Xue, Constant, Roberts, Kale, Al-Rfou, Siddhant,
  Barua, and Raffel}]{xue2020mt5}
Linting Xue, Noah Constant, Adam Roberts, Mihir Kale, Rami Al-Rfou, Aditya
  Siddhant, Aditya Barua, and Colin Raffel. 2020.
\newblock mt5: A massively multilingual pre-trained text-to-text transformer.
\newblock \emph{arXiv preprint arXiv:2010.11934}.

\bibitem[{Yang et~al.(2023)Yang, Xiao, Wang, Zhang, Bian, Yin, Lv, Pan, Wang,
  Yan, Yang, Deng, Wang, Liu, Ai, Dong, Zhao, Xu, Sun, Zhang, Liu, Ji, Xie,
  Dai, Fang, Su, Song, Liu, Ru, Ma, Wang, Liu, Lin, Nie, Guo, Sun, Zhang, Li,
  Li, Cheng, Chen, Zeng, Wang, Chen, Men, Yu, Pan, Shen, Wang, Li, Jiang, Gao,
  Zhang, Zhou, and Wu}]{yang2023baichuan}
Aiyuan Yang, Bin Xiao, Bingning Wang, Borong Zhang, Ce~Bian, Chao Yin, Chenxu
  Lv, Da~Pan, Dian Wang, Dong Yan, Fan Yang, Fei Deng, Feng Wang, Feng Liu,
  Guangwei Ai, Guosheng Dong, Haizhou Zhao, Hang Xu, Haoze Sun, Hongda Zhang,
  Hui Liu, Jiaming Ji, Jian Xie, JunTao Dai, Kun Fang, Lei Su, Liang Song,
  Lifeng Liu, Liyun Ru, Luyao Ma, Mang Wang, Mickel Liu, MingAn Lin, Nuolan
  Nie, Peidong Guo, Ruiyang Sun, Tao Zhang, Tianpeng Li, Tianyu Li, Wei Cheng,
  Weipeng Chen, Xiangrong Zeng, Xiaochuan Wang, Xiaoxi Chen, Xin Men, Xin Yu,
  Xuehai Pan, Yanjun Shen, Yiding Wang, Yiyu Li, Youxin Jiang, Yuchen Gao,
  Yupeng Zhang, Zenan Zhou, and Zhiying Wu. 2023.
\newblock \href {http://arxiv.org/abs/2309.10305} {Baichuan 2: Open large-scale
  language models}.

\bibitem[{Zellers et~al.(2019)Zellers, Holtzman, Bisk, Farhadi, and
  Choi}]{zellers2019hellaswag}
Rowan Zellers, Ari Holtzman, Yonatan Bisk, Ali Farhadi, and Yejin Choi. 2019.
\newblock Hellaswag: Can a machine really finish your sentence?
\newblock \emph{arXiv preprint arXiv:1905.07830}.

\bibitem[{Zeng et~al.(2022)Zeng, Liu, Du, Wang, Lai, Ding, Yang, Xu, Zheng, Xia
  et~al.}]{zeng2022glm}
Aohan Zeng, Xiao Liu, Zhengxiao Du, Zihan Wang, Hanyu Lai, Ming Ding, Zhuoyi
  Yang, Yifan Xu, Wendi Zheng, Xiao Xia, et~al. 2022.
\newblock Glm-130b: An open bilingual pre-trained model.
\newblock \emph{arXiv preprint arXiv:2210.02414}.

\bibitem[{Zhou et~al.(2023)Zhou, Li, Jiang, and Bing}]{zhou2023enhancing}
Meng Zhou, Xin Li, Yue Jiang, and Lidong Bing. 2023.
\newblock Enhancing cross-lingual prompting with dual prompt augmentation.
\newblock In \emph{Findings of the Association for Computational Linguistics:
  ACL 2023}, pages 11008--11020.

\bibitem[{Ács(2019)}]{fertility}
Judit Ács. 2019.
\newblock \href
  {http://juditacs.github.io/2019/02/19/bert-tokenization-stats.html}
  {Exploring bert's vocabulary}.

\end{thebibliography}

\label{lr:ref}
\bibliographystylelanguageresource{lrec-coling2024-natbib}
\bibliographylanguageresource{languageresource}

\section{Appendix}

\subsection{KoBEST Evaluation Results}\label{a:kobest}

\begin{table}[b]
\resizebox{\columnwidth}{!}{%
\begin{tabular}{lrrrrr}
\hline
Model & \multicolumn{1}{l}{Lang} & \multicolumn{1}{l}{BoolQ} & \multicolumn{1}{l}{COPA} & \multicolumn{1}{l}{HellaSwag} & \multicolumn{1}{l}{SentiNeg} \\ \hline
\multirow{2}{*}{GPT-3.5-Turbo} & Ko & 82.34 & 57.88 & 40.00 & 91.69 \\
 & En & 86.40 & 79.20 & 55.00 & 96.73 \\
\multirow{2}{*}{GPT-4} & Ko & 96.58 & 52.40 & 76.60 & 98.74 \\
 & En & 96.65 & 95.70 & 73.00 & 98.49 \\ \hline
\end{tabular}%
}
\caption{Evaluation results for GPT-3.5 and GPT-4 on KoBEST}
\label{a:gptkb}
\end{table}

\begin{table*}[htbp]
\resizebox{\textwidth}{!}{%
\begin{tabular}{lccccccc}
\hline
Model & Lang & \begin{tabular}[c]{@{}c@{}}Loan \\ Words\end{tabular} & \begin{tabular}[c]{@{}c@{}}Standard\\ Nomenclature\end{tabular} & \begin{tabular}[c]{@{}c@{}}Rare\\ Words\end{tabular} & History & \begin{tabular}[c]{@{}c@{}}General \\ Knowledge\end{tabular} & \begin{tabular}[c]{@{}c@{}}Reading\\ Comprehension\end{tabular} \\ \hline
\multirow{2}{*}{GPT-3.5-Turbo} & Ko & 62.13 & 55.56 & 63.46 & 30.32 & 35.80 & 60.18 \\
 & En & \textbf{72.19} & 67.32 & 61.73 & 30.85 & 42.61 & 57.72 \\
\multirow{2}{*}{GPT-4} & Ko & 70.41 & 67.32 & \textbf{74.32} & \textbf{60.64} & 54.55 & \textbf{79.64} \\
 & En & 66.86 & \textbf{79.08} & 73.83 & 54.79 & \textbf{55.68} & 79.19 \\ \hline
\end{tabular}%
}
\caption{Evaluation results for GPT-3.5 and GPT-4 on HAE-RAE Bench}
\label{a:gptHR}
\end{table*}

\begin{table*}[t]
\resizebox{\textwidth}{!}{%
\begin{tabular}{lrrrrcrrrcrrrlrrr}
\hline
 & \multicolumn{1}{l}{} & \multicolumn{3}{c}{BoolQ} & \multicolumn{1}{l}{} & \multicolumn{3}{c}{COPA} & \multicolumn{1}{l}{} & \multicolumn{3}{c}{HellaSwag} &  & \multicolumn{3}{c}{SentiNeg} \\ \cline{3-5} \cline{7-9} \cline{11-13} \cline{15-17} 
Model & \multicolumn{1}{l}{Params} & \multicolumn{1}{c}{n=0} & \multicolumn{1}{c}{n=5} & \multicolumn{1}{c}{n=10} &  & \multicolumn{1}{c}{n=0} & \multicolumn{1}{c}{n=5} & \multicolumn{1}{c}{n=10} &  & \multicolumn{1}{c}{n=0} & \multicolumn{1}{c}{n=5} & \multicolumn{1}{c}{n=10} &  & \multicolumn{1}{l}{n=0} & \multicolumn{1}{l}{n=5} & \multicolumn{1}{l}{n=10} \\ \hline
\multirow{4}{*}{Polyglot-Ko} & 1.3B & 49.9 & 51.2 & 50.2 &  & 72.1 & 72.0 & 71.8 &  & 41.0 & 40.6 & 42.0 &  & 69.8 & 62.0 & 56.2 \\
 & 3.8B & 50.6 & 53.9 & 52.9 &  & 75.7 & 76.2 & 76.3 &  & 44.6 & 47.8 & 48.0 &  & 58.7 & 81.1 & 76.1 \\
 & 5.8B & 53.7 & 58.6 & 55.7 &  & 77.9 & 76.9 & 77.5 &  & 49.0 & 48.6 & 50.2 &  & 50.4 & 87.9 & 88.4 \\
 & 12.8B & \textbf{56.7} & 62.8 & 63.8 &  & \textbf{79.6} & \textbf{81.0} & \textbf{80.2} &  & \textbf{49.0} & \textbf{51.0} & \textbf{49.8} &  & \textbf{91.7} & \textbf{90.7} & \textbf{93.5} \\
\multirow{2}{*}{UMT5} & 3B & 50.2 & 50.2 & 50.2 &  & 52.4 & 53.4 & 51.1 &  & 32.0 & 31.0 & 28.2 &  & 52.4 & 49.9 & 49.6 \\
 & 13B & 50.2 & 50.3 & 50.3 &  & 57.9 & 58.2 & 57.3 &  & 36.2 & 32.0 & 31.6 &  & 56.9 & 51.9 & 46.4 \\
\multirow{2}{*}{Llama-2} & 7B & 50.9 & 58.8 & 55.8 &  & 56.1 & 58.1 & 58.2 &  & 41.8 & 43.2 & 43.2 &  & 48.9 & 58.2 & 57.2 \\
 & 13B & 50.6 & \textbf{74.2} & \textbf{77.1} &  & 59.1 & 64.1 & 63.1 &  & 41.6 & 44.8 & 42.6 &  & 50.6 & 73.8 & 85.9 \\ \hline
\end{tabular}%
}
\caption{Evaluation results for KoBEST.}
\label{kobest}
\end{table*}

The evaluation results for KoBEST are presented in table~\ref{kobest}. Polyglot-Ko achieves the top scores in most settings, with exceptions in the 5-shot and 10-shot configurations for BoolQ.

\subsection{GPT-3.5/4 Evaluation}\label{a:gpt}

In this section, we detail the prompts employed for the XLT evaluation of GPT-3.5 and GPT-4 models. We also provide their performance metrics for each downstream task in the HAE-RAE Bench and KoBEST datasets. The specific instructions used for XLT with both models are presented below. Table~\ref{a:gptHR} and Table~\ref{a:gptkb} contain the evaluation results for HAE-RAE Bench and KoBEST, respectively.

\begin{figure}[h!]
\fontsize{8}{9}\selectfont
    \centering
    \begin{tabular}{|p{0.9\linewidth}|}
    \hline 
    \\
    Read the given question, and choose the most suitable answer. Answer your answer with the number of the answer you think to be correct. \\
    \#\#\# question: \{question\} \\
    \#\#\# options: \\(1) \{option\#1\} \\(2) \{option\#3\} \\(3) \{option\#3\} \\(4) \{option\#4\} \\(5) \{option\#5\} \\
    \#\#\# answer:
\\\\
\hline
    \end{tabular}
    \caption{Prompt used in our Direct Evaluation.}
    \label{fig:direct-prompt}
\end{figure}

\subsection{HAE-RAE Bench Examples}\label{a:ex}

\begin{figure*}
  \centering
  \includegraphics[width=\textwidth]{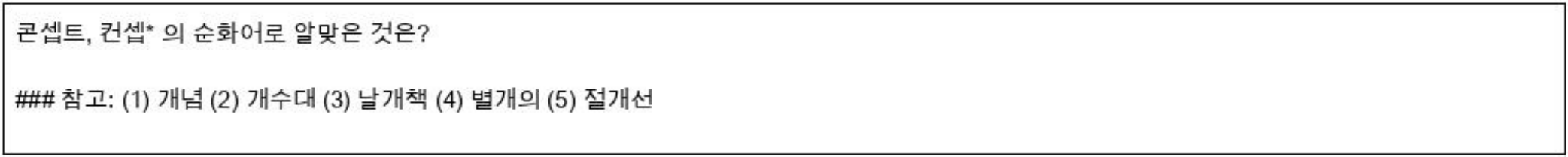}
  \caption{An example from the Loan Words subset.}
  \label{lwko}
\end{figure*}

Starting from Figure~\ref{lwko}, we present examples for each task alongside their translated versions.

\begin{figure*}
  \centering
  \includegraphics[width=\textwidth]{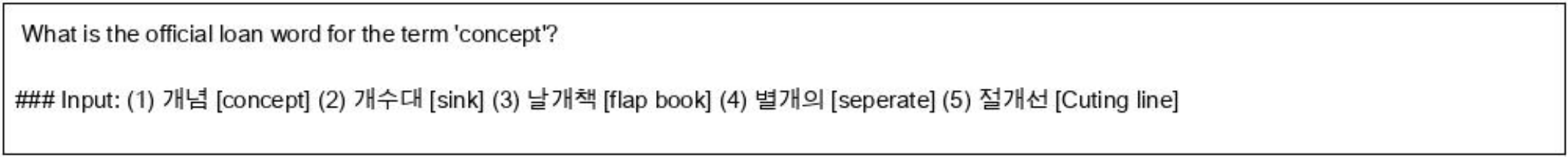}
  \caption{A translated example from the Loan Words subset.}
\end{figure*}

\begin{figure*}
  \centering
  \includegraphics[width=\textwidth]{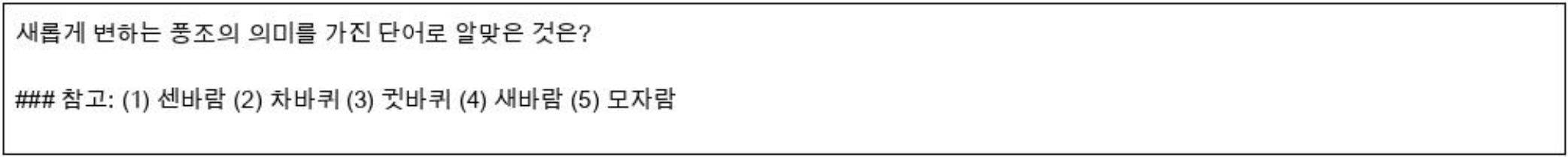}
  \caption{An example from the Rare Words subset.}
\end{figure*}

\begin{figure*}
  \centering
  \includegraphics[width=\textwidth]{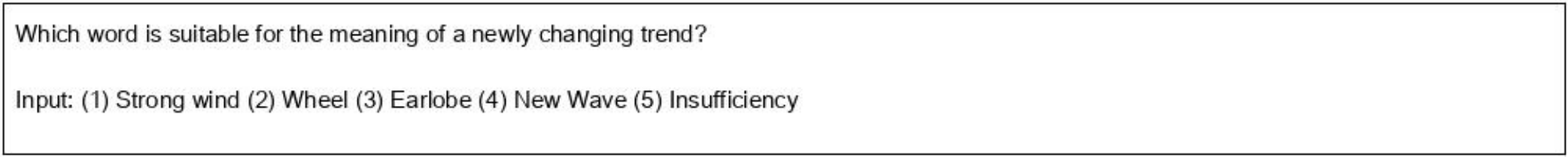}
  \caption{A translated example from the Rare Words subset.}
\end{figure*}

\begin{figure*}
  \centering
  \includegraphics[width=\textwidth]{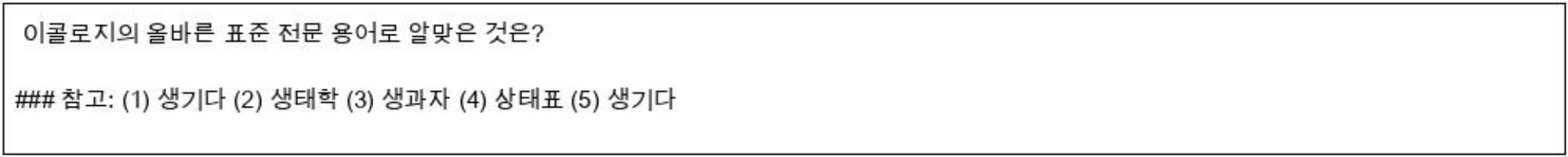}
  \caption{An example from the Standard Nomenclature subset.}
\end{figure*}

\begin{figure*}
  \centering
  \includegraphics[width=\textwidth]{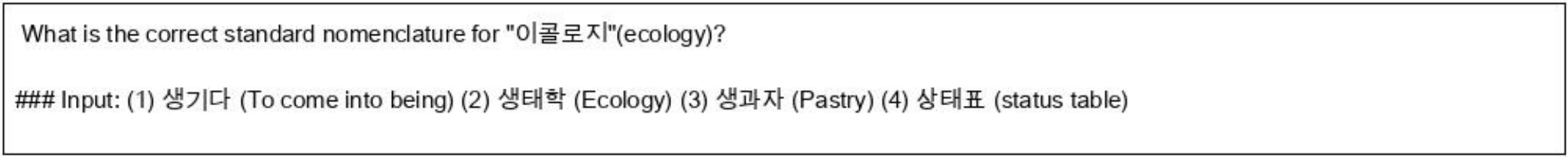}
  \caption{A translated example from the Standard Nomenclature subset.}
\end{figure*}

\begin{figure*}
  \centering
  \includegraphics[width=\textwidth]{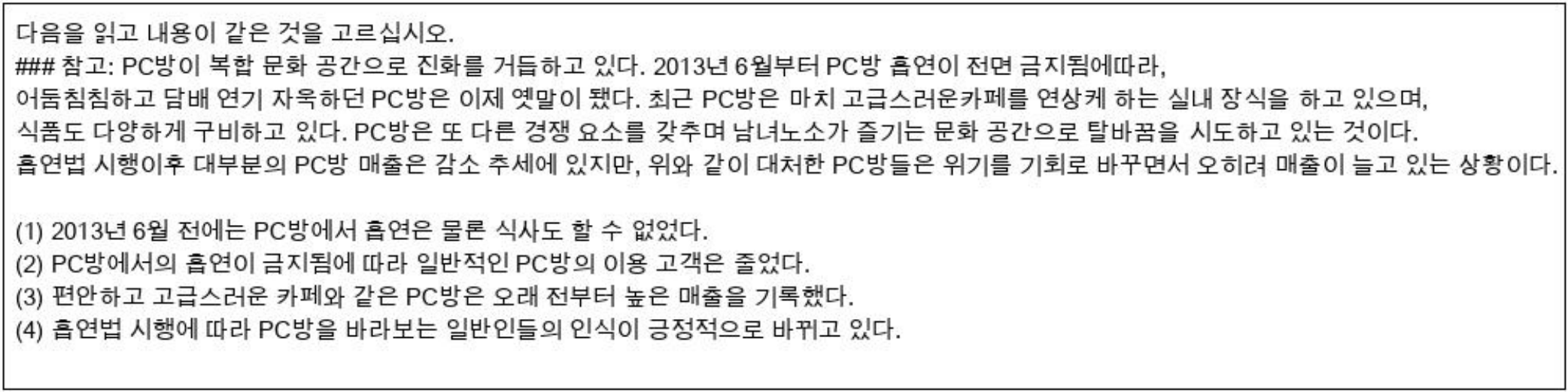}
  \caption{An example from the Reading Comprehension subset.}
\end{figure*}

\begin{figure*}
  \centering
  \includegraphics[width=\textwidth]{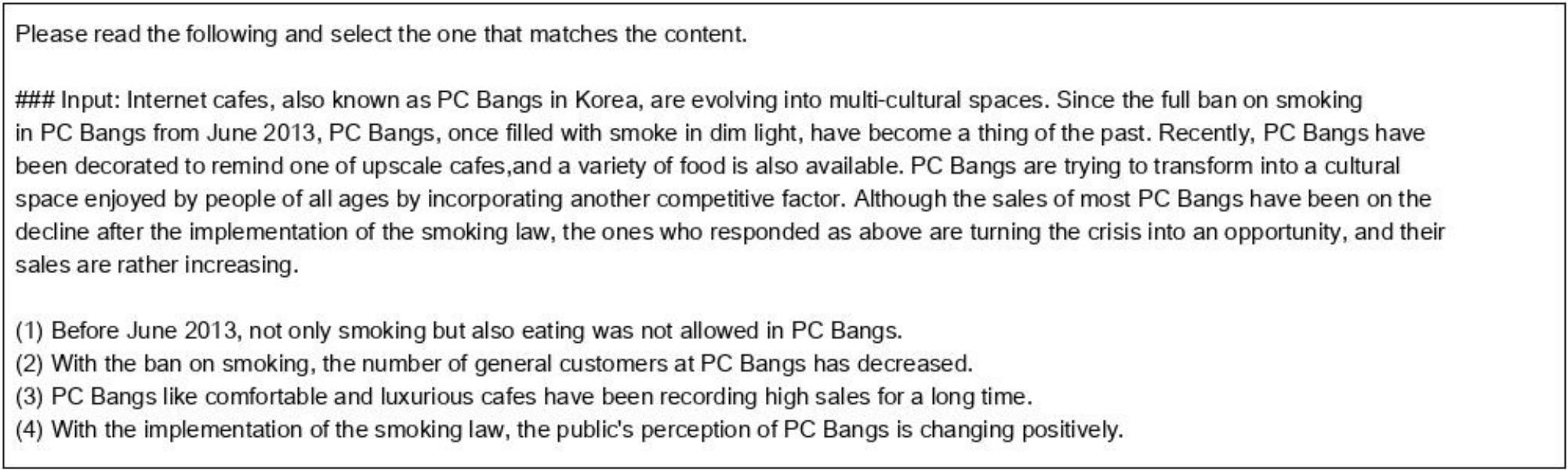}
  \caption{A translated example from the Reading Comprehension subset.}
\end{figure*}

\begin{figure*}
  \centering
  \includegraphics[width=\textwidth]{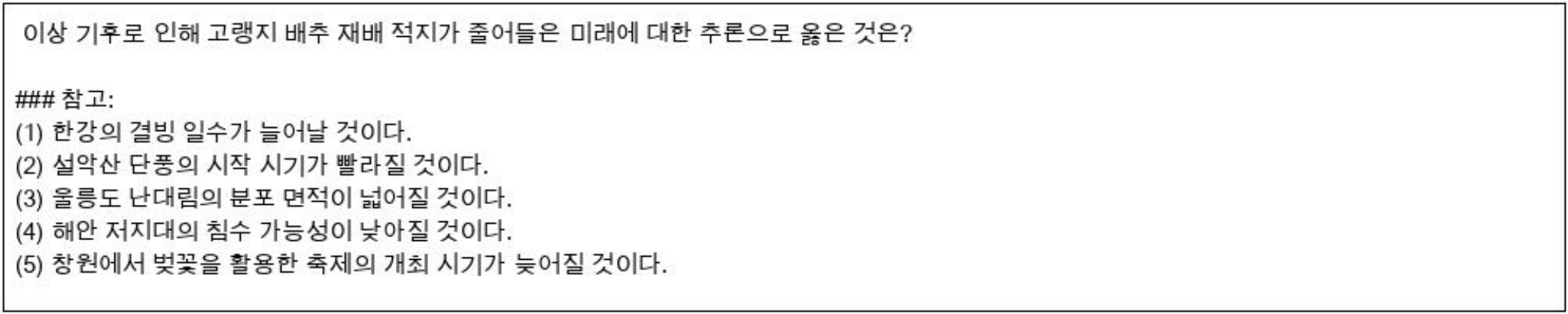}
  \caption{An example from the General Knowledge subset.}
\end{figure*}

\begin{figure*}
  \centering
  \includegraphics[width=\textwidth]{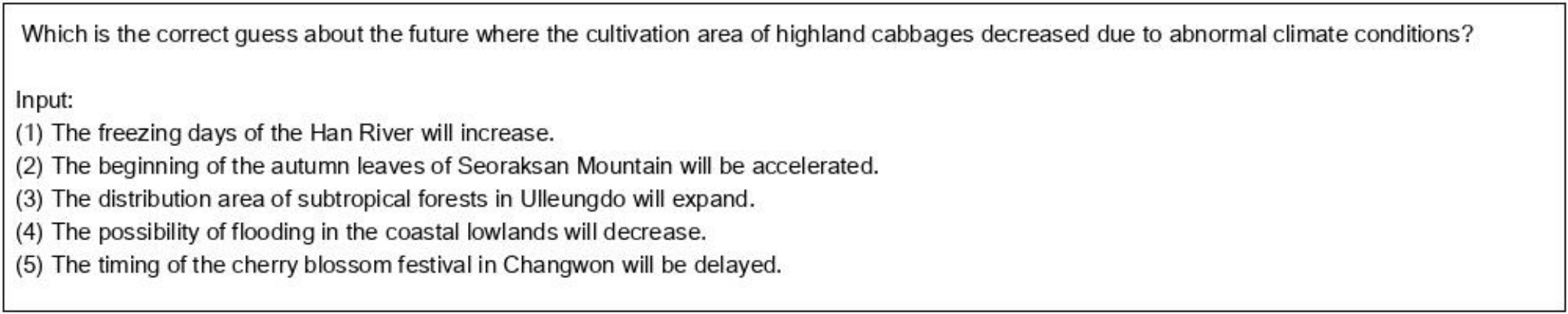}
  \caption{A translated example from the General Knowledge subset.}
\end{figure*}

\begin{figure*}
  \centering
  \includegraphics[width=\textwidth]{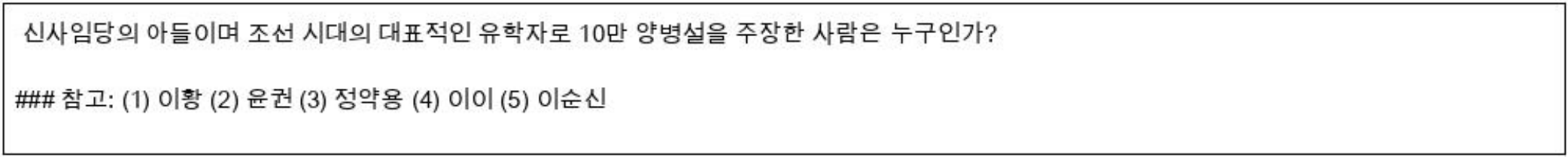}
  \caption{An example from the History subset.}
\end{figure*}

\begin{figure*}
  \centering
  \includegraphics[width=\textwidth]{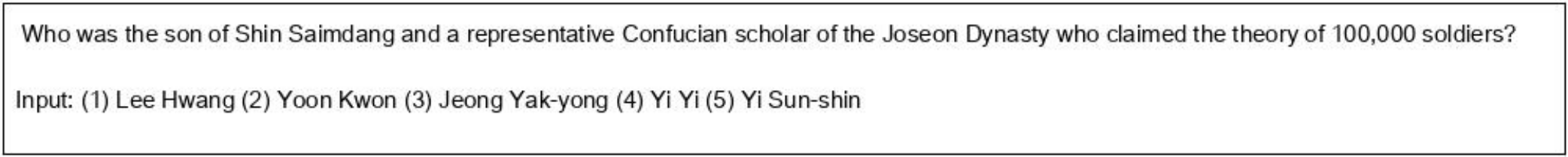}
  \caption{A translated example from the History subset.}
\end{figure*}

\end{document}